\documentclass{article}

    \usepackage[final]{neurips_2023}

\usepackage[utf8]{inputenc} 
\usepackage[T1]{fontenc}    
\usepackage{hyperref}       
\usepackage{url}            
\usepackage{booktabs}       
\usepackage{amsfonts}       
\usepackage{nicefrac}       
\usepackage{microtype}      
\usepackage{xcolor}         
\usepackage{amsmath}
\usepackage{graphicx}
\usepackage{subcaption}
\usepackage{natbib}
\usepackage{multirow}
\usepackage{mathrsfs}
\usepackage{bbm}
\usepackage{xr}
\setcitestyle{authoryear,round, semicolon}

\title{Non-parametric Probabilistic Time Series Forecasting via Innovations Representation}

\author{%
  Xinyi Wang \\
  Cornell University\\
  Ithaca, NY 14850 \\
  \texttt{xw555@cornell.edu} \\
  \And
  Mei-jen Lee\\
  Cornell University\\
  Ithaca, NY 14850 \\
  \texttt{ml2298@cornell.edu} \\
  \And
  Qing Zhao\\
    Cornell University\\
  Ithaca, NY 14850 \\
  \texttt{qz16@cornell.edu} \\
  \And
  Lang Tong\\
    Cornell University\\
  Ithaca, NY 14850 \\
  \texttt{lt35@cornell.edu} \\
}

\newtheorem{theorem}{Theorem}

\newtheorem{definition}{Definition}

\def\beq{\begin{equation}}
\def\eeq{\end{equation}}
\def\bea{\begin{eqnarray}}
\def\eea{\end{eqnarray}}
\def\ba{\begin{array}}
\def\ea{\end{array}}

\def\bitem{\begin{itemize}}
\def\eitem{\end{itemize}}
\def\ben{\begin{enumerate}}
\def\een{\end{enumerate}}

\definecolor{bgrd}{rgb}{1,1,1}
\definecolor{gray}{rgb}{0.5,0.5,0.5}
\definecolor{dkr}{rgb}{0.7,0.1,0.2}
\definecolor{dkb}{rgb}{0.1,0.1,0.8}

\makeatletter
\newdimen{\captionwidth}
\long\def\@makecaption#1#2{%
\captionwidth .9\hsize
\vskip 10pt%
\setbox\@tempboxa\hbox{#1: #2}%
  \ifdim \wd\@tempboxa >\captionwidth%
    \setbox\@tempboxa\hbox{#1:\hspace*{.5em}}%
    \hfil\parbox{\captionwidth}{\raggedright\hangindent \wd\@tempboxa%
    \hangafter=1\unhbox\@tempboxa#2}\hfill%
  \else\centerline{\box\@tempboxa}%
  \fi
}
\makeatother

\newcommand{\mbbE}{\mathbb {E}}

\newcommand{\Xmsc}{\mathscr{X}}

\def\nubf{\hbox{\boldmath$\nu$\unboldmath}}

\def\Ac{{\cal A}}

\def\Uc{{\cal U}}

\begin{document}

\maketitle
\begin{abstract}
Probabilistic time series forecasting predicts the conditional probability distributions of the time series at a future time given past realizations. Such techniques are critical in risk-based decision-making and planning under uncertainties. 
Existing approaches are primarily based on parametric or semi-parametric time-series models that are restrictive, difficult to validate, and challenging to adapt to varying conditions. This paper proposes a nonparametric method based on the classic notion of {\em innovations} pioneered by Norbert Wiener and Gopinath Kallianpur that causally transforms a nonparametric random process to an independent and identical uniformly distributed {\em innovations process}. 
We present a machine-learning architecture and a learning algorithm that circumvent two limitations of the original Wiener-Kallianpur innovations representation: (i) the need for known probability distributions of the time series and (ii) the existence of a causal decoder that reproduces the original time series from the innovations representation.   
We develop a deep-learning approach and a Monte Carlo sampling technique to obtain a generative model for the predicted conditional probability distribution of the time series based on a weak notion of Wiener-Kallianpur innovations representation. 
The efficacy of the proposed probabilistic forecasting technique is demonstrated on a variety of electricity price datasets, showing marked improvement over leading benchmarks of probabilistic forecasting techniques.  
\end{abstract}

\section{Introduction}
\label{sec:intro}
We consider the problem of {\em nonparametric probabilistic time-series forecasting} that predicts the probability distribution of a random process at a future time instance, conditional on current and past realizations. Whereas {\em point forecasting} that predicts a specific future realization of the time series represents most forecasting needs, probabilistic forecasting is critical in applications where the decision-making process must incorporate probability distributions that characterize future uncertainty.
With probabilistic forecasting, point forecasting can also be derived as a by-product.

Probabilistic time-series forecasting is challenging because probabilistic time-series models are infinite-dimensional, unknown, and with complex temporal dependencies and joint probability distributions. Conventional approaches are mostly parametric and semiparametric that reduce the infinite-dimensional inference problem to one within a finite-dimensional parameter space. Classic examples include the forecasting based on autoregressive moving average, GARCH, and Gaussian process models \citep{weron_forecasting_2008,weron_electricity_2014}.

This work pursues a new path to nonparametric probabilistic forecasting based on the notion of {\em innovations representation} pioneered by Norbert Wiener and Gopinath Kallianpur in 1958 \citep{Wiener:58Book}. 
The Wiener-Kalliapur innovations representation is defined by a {\em causal autoencoder}, where the encoder transforms a stationary random process to the canonical form of an {\em innovations process}---an independent and identical uniformly distributed (i.i.d. uniform) sequence---and a causal decoder that reproduces the original. The notion of innovations arises from that, as a latent variable of the autoencoder,  the innovations process at time $t$ captures the new information of the input process at time $t$ statistically independent of its past.  
The Wiener-Kallianpur innovations representation, when it exists, provides powerful means for real-time decision-making because it reduces a general stationary process to the the canonical i.i.d. uniform process, on which decisions can be made with no loss. Classic examples of Wiener filtering, Kalman filtering, and linear quadratic Gaussian control are instances of applying Wiener-Kallianpur innovations representation to Gaussian processes with known distributions \citep{bertsekas_dynamic_2012,kailath_linear_2000}. 

The main challenge of applying Wiener-Kalliapur innovations representation to inference and decision-making problems is twofold. 
First, obtaining a causal encoder to extract the innovations process requires knowing the marginal and joint distributions of the time series, which is rarely possible without imposing some parametric structure on the time series. Furthermore, even when the probability distribution is known, there is no known computationally tractable way to construct the causal encoder that transforms the time series to the innovations process. Second, it turns out that Wiener-Kalliapur innovations representation may not exist for a broad class of random processes, including some of the important cases of finite-state Markov chains \citep{Rosenblatt:59}. These conceptual and computational barriers prevent employing innovations representation to the broad class of inference and decision problems except under the special Gaussian and additive Gaussian models \citep{Kailath1968TAC}. 
\vspace{-2em}
\subsection{Summary of Contributions}
\vspace{-0.5em}
This paper makes methodological and practical advances in nonparametric probabilistic time series forecasting in three aspects.
First, we develop a GAN-based deep-learning architecture and a machine-learning algorithm to obtain a weak version of the Wiener-Kallianpur innovations sequence by relaxing the requirement that the original time series can be perfectly reconstructed from the innovations process by a causal decoder. In particular, we propose a Weak Innovations AutoEncoder (WIAE) consisting of a causal encoder, a {\em weak innovations sequence} as the latent random process, and a causal decoder whose output has matching distributions with the WIAE input.   
Under WIAE, the weak innovations process has the same form and interpretation as that of Wiener-Kallianpur, i.e., the innovation at time $t$ contains only new information not present in past samples. However, the innovations extracted under WIAE does not guarantee to reproduce the original time series with the causal decoder. The significance of WIAE is that it eliminates the need to know the underlying joint probability distributions of the time series and greatly enlarges the class of time series for which the canonical uniform i.i.d. innovations representation exists. Since practical implementations of forecasting cannot involve samples from the infinite past, we establish convergence results in Theorem.~\ref{thm:converge}, showing that, with increasingly higher dimensionalities of the auto-encoder, optimally trained finite dimensional WIAE converges to the WIAE limit. See Sec.~\ref{subsec:training}.

Second, we argue that, despite that the original time series are not reproducible from the weak innovations sequence, the conditional distribution of the time series at a future time given past realizations is identical to the conditional distribution given past samples of the innovations process. This implies that, for the probabilistic time series forecasting problem, there is no loss of using the innovations rather than the original time series in forecasting. The significance of this equivalence is that probabilistic forecasting of the future requires joint probability distributions of yet realized future random variables. When innovations sequence is used in forecasting, the joint probability distribution of all future innovations is known to be i.i.d. uniform, which makes it possible to obtain simple Monte Caro samples for forecasting. This insight leads to a novel solution to nonparametric probabilistic time series forecasting. See Sec.~\ref{subsec:inference}.

Third, we conducted extensive empirical evaluations based on public datasets, comparing the proposed innovations-based forecasting with parametric and nonparametric techniques in classic statistics and leading machine learning benchmarks. To this end, we focus on probabilistic forecasting of real-time electricity prices, which is one of the most challenging forecasting tasks because electricity prices are dual variables of constrained optimization highly sensitive to stochastic demand and environmental factors. Price forecasting is also highly desirable by a system operator for scheduling generation and demand under uncertainty and by market participants to construct profit-maximizing bids and offers. The empirical results demonstrate marked improvement over existing solutions, scoring near the top across multiple datasets and under different performance metrics.   
See Sec.~\ref{sec:simulation}.

\subsection{A Contextual Review of Literature}
\label{subsec:literature}
The literature on time series forecasting is vast but somewhat limited for nonparametric probabilistic forecasting solutions. Here we restrict our review to probabilistic and related point forecasting techniques that share architectural or methodological similarities with the proposed solution.

\cite{hardle_review_1997} give a review of classic nonparametric time series analysis, where the starting point is a nonlinear regression model driven by an innovations process. These neoclassic techniques aimed to obtain the conditional distributions directly rather than obtaining generative models \citep{tong_threshold_1983,haggan_modelling_1981,chan_estimating_1986,granger_modelling_1993, hart_automated_1996,robinson_nonparametric_1983,auestad_identification_1990,hardle_kernel_1992,tjostheim_nonparametric_1994,hardle_nonparametric_1998}.
A primary difference between these neoclassic statistical techniques and modern deep learning approaches is that kernels and forms of these estimators are chosen a priori rather than learned from data. In spite of their principled methodology with some level of performance guarantees \citep{robinson_nonparametric_1983,masry_nonparametric_1995}, they tend to suffer from difficulties in practical implementation and the "curse of dimensionality". 

Modern machine learning ushers novel learning architectures and learning algorithms such as deep recurrent neural networks (RNN) with long short-term memory (LSTM) cells capable of modeling more complex temporal dependencies \citep{salinas_deepar_2019,salinas_high-dimensional_2019,rangapuram_deep_2018,wang_deep_2019,yoon_robust_2022,du_probabilistic_2022}, which led to a variety of parametric and semiparametric probabilistic forecasting techniques. 
\cite{oord_wavenet_2016,borovykh_conditional_2018} adopted auto-regressive neural network with dilated convolutional network to learn the conditional distribution. 
Most relevant to the innovations representation approach proposed in this work is using the autoencoder architecture for probabilistic forecasting.   In particular, \cite{nguyen_temporal_2021} developed a probabilistic forecasting technique that models the predicted conditional distribution as a Gaussian latent process of an autoencoder trained with historical data. \cite{li_synergetic_2021} adopted a conditional variational auto-encoder to learn the latent variable (typically of Gaussian distribution) for asynchronous temporal point processes. The decoder was then used to produce probabilistic forecasts. \cite{le_guen_probabilistic_2020} used an autoencoder and a diversification mechanism relying on determinantal point process to enhance the performance of forecasting for non-stationary time series. 

The success of deep learning brought a wave of point forecast techniques based on deep advanced deep learning architecture.
Gaining much recent attention are those methods based on the idea of {\em transformer} derived from natural language processing \citep{lim_temporal_2020,zhou_informer_2021,zhou_fedformer_2022,liu_pyraformer_2022,zhang_crossformer_2023,nie_time_2023,zhou_fedformer_2022}, demonstrating promising performance for certain types of datasets. Although these techniques do not produce probabilistic forecasts, their strong performance in point forecasting warrants a comparison study with the proposed probabilistic technique specialized for the point forecasting task.

This work is inspired by the seminal work of innovations representation of Wiener and Kallianpur \citep{Wiener:58Book}.  Having discovered that the Wiener-Kallianpur innovations representation may not exist in general, the idea of weak innovations representation was suggested by \cite{Rosenblatt:59}, who proposed to relax the requirement of perfect reproduction of the original time series to that the output of WIAE matches the input in distribution.
Our work focuses on a machine learning approach to extract weak innovations for forecasting.  To this end, we adopt the deep learning architecture of Wang and Tong (2022) designed to extract Wiener-Kallianpur innovations process with different learning objectives, learning algorithm, and convergence properties.  The use of innovations for time series forecasting has not been considered in the past, and the techniques presented in this work is novel.

\vspace{-1em}
\section{Innovations Representation}
\vspace{-0.5em}
\label{sec:innovations}
\label{subsec:inn}
The Wiener-Kallianpur Innovations (Wiener 1958) representation of a time series $(x_t)$ is defined by an autoencoder with a pair of causal transforms $(G,H)$ and a latent innovations process $(\nu_t)$ in the form of i.i.d uniform sequence, where
\begin{align}
    \nu_t &= G(x_t,x_{t-1},\cdots),\label{eq:s-encoder}\\
    x_t &= H(\nu_t,\nu_{t-1},\cdots),\label{eq:s-decoder}\\
    &\nu_t\stackrel{\mbox{\sf\tiny i.i.d}}{\sim}\Uc[0,1]\label{eq:s-iid}
\end{align}
Note that Eq.~\eqref{eq:s-encoder} and \eqref{eq:s-iid} imply that $\nu_t$ is independent of $\Xmsc_{t-1}:=\{x_{t-1},x_{t-2},\cdots\}$, giving the interpretation that $\nu_t$ represents the new information not in $\Xmsc_{t-1}$.

The Wiener-Kallianpur innovations representation originated from the problem to encode in the most efficient fashion a general stationary random process \citep{Wiener:58Book}. To that end, it is natural to require that the decoder reproduce the input, and the i.i.d. uniform innovations appear to be the simplest and the most efficient (with the highest entropy)  representation. Later, \cite{Kailath1968TAC} shows that, under the Gaussian and additive Gaussian assumptions, innovations process can be extracted by a least squares predictor, which appears to be the only known way of obtaining the causal autoencoder for nonparametric Gaussian processes.  When the random process has a state-space model, the celebrated Kalman filtering \citep{Kalman:60TASME} is another instance of extracting innovations process from the observation process.
Besides these special cases, there are no general technique to obtain the Wiener-Kallianpur innovations for general non-parametric random processes, until recently when \cite{WangTong:21JMLR} obtained the first data-driven technique to extract innovations sequence via a causal auto-encoder trained with Generative Adversarial Networks (GANs).  

\cite{Rosenblatt:59} was the first to point out that the Wiener-Kallianpur innovations may not exist for a broad class of stationary random processes.  He conjectured a weaker version of the innovations representation may be of interest.  
Inspired by that observation, introduced here is the {\em weak innovations representation} defined by a causal WIAE by
\begin{align}
    \nu_t &= G(x_t,x_{t-1},\cdots), \label{Eq:encoder}\\
    \hat{x}_t &= H(\nu_t,\nu_{t-1},\cdots), \label{Eq:decoder}\\
    &\nu_t\stackrel{\mbox{\sf\tiny i.i.d}}{\sim} \Uc[0,1],\label{Eq:iid}\\
    &(\hat{x}_t) \stackrel{d}{=} (x_t) \label{Eq:recons},
\end{align}
The weak innovations sequence is defined by relaxing the invertible function pair defined for innovations representation, to "invertible by distribution" defined by Eq.~(\ref{Eq:encoder}\&\ref{Eq:decoder}).
For probabilistic time series forecasting problems, only information with respect to the conditional distribution of $(x_t)$ is of interest.
Therefore, intuitively, the representation $(\nu_t)$'s capability of reconstructing a sequence $(\hat{x}_t)$ with the same distribution\footnote{This means that for any finite set of time indices $\mathcal{T}$, the joint distribution of $(x_t)_{t\in\mathcal{T}}$ is the same as the joint distribution of $(\hat{x}_t)_{t\in\mathcal{T}}$.} as $(x_t)$ should is sufficient for time series forecasting problems.
Since the weak innovations representation is calculated causally, it is also suitable to be implemented for any real-time applications.

Currently, there is no known techniques to extract weak innovations sequence from time series.
We believe that we made the first attempt to solve open-ended question of extracting and leveraging the weak innovations sequence.

\section{WIAE: Training and Probablistic Forecasting}
\label{sec:wiae}
\begin{figure}
\begin{subfigure}[b]{0.45\linewidth}
    \centering
    \includegraphics[width=\linewidth]{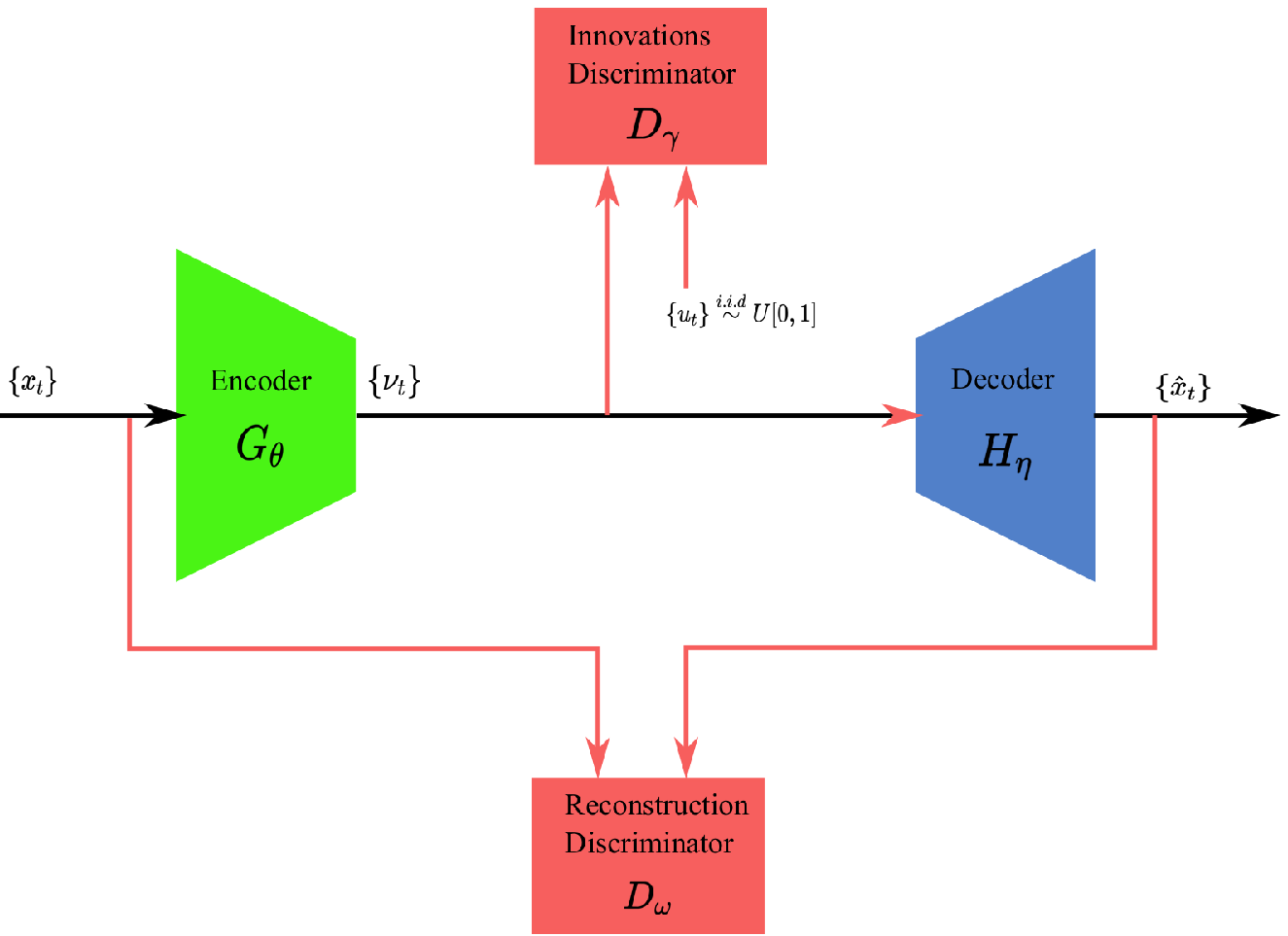}
    \caption{Training}
    \label{fig:training scheme}
\end{subfigure}
\begin{subfigure}[b]{0.45\linewidth}
    \centering
    \includegraphics[width=\linewidth]{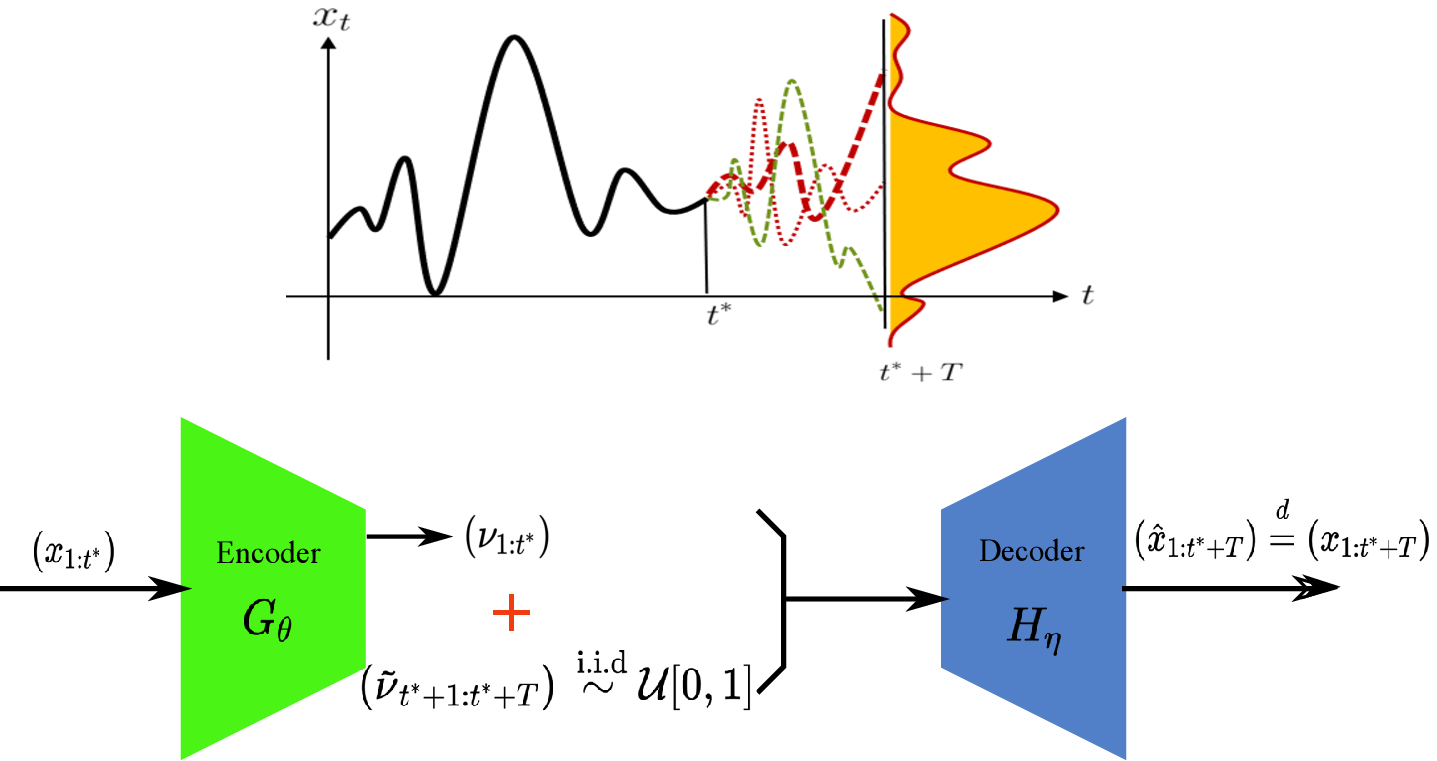}
    \caption{$T$-step forecast scheme via WIAE. By repeated sampling i.i.d innovations, one can get the empirical joint distribution of $(x_{t^*+1:t^*+T})$ represented by the trajectories.}
    \label{fig:prediction}
\end{subfigure}
\caption{Schematics}
\label{fig:scheme}
\end{figure}
This section we introduce the schematics and training algorithm to obtain weak innovations representation, and develop a probabilistic forecasting solution.
As shown in Fig.~\ref{fig:scheme}, WIAE is a causal autoencoder that generates the weak innovations process in the forms of i.i.d. uniform sequence as defined in (4-7). In practice, $(G, H)$ are approximated by finite dimensional implementations $(G_{\theta_m}, H_{\eta_m})$ with input dimension\footnote{A pair of auto-encoder with dimension $m$ means that its inputs only consist of the past $m$ samples: $(x_t,x_{t-1},\cdots,x_{t-m+1})$.} $m$, constructed from causal time-delay convolutional layers \citep{Waibel&etal:89}.  
Sequences used for training are reformatted to vectors for compactness, hence the input sequence $(x_t)$ being reshaped as vectors in the sliding window fashion.
The vector version is denoted through boldface notations, i.e., $\boldsymbol{G}^{(n)}_{\theta_m}$ and $\boldsymbol{H}^{(n)}_{\eta_m}$. 
The superscripts $n$ represents the dimension of input to the discriminators, i.e., the input to vectored generator $\boldsymbol{G}^{(n)}_{\theta_m}$ is $[[x_{1},x_{2},\cdots,x_m],[x_{m+1},x_{m+2}\cdots,x_{2m}],\cdots,[x_{1+(n-1)m},x_{2+(n-1)m},\cdots,x_{nm}]]$. 
The outputs of $\boldsymbol{G}^{(n)}_{\theta_m}$ and $\boldsymbol{H}^{(n)}_{\eta_m}$ were denoted by $\boldsymbol{\nu}_{t,m}^{(n)}$ and $\hat{\boldsymbol{x}}_{t,m}^{(n)}$. 

\subsection{Training and Structural Convergence}
\label{subsec:training}
To learn the transformation pair $(G,H)$, the latent representation learned by auto-encoder pair ${G_\theta,H_\eta}$ needs to be constrained according to Eq.~\eqref{Eq:iid} \& Eq.~\eqref{Eq:recons}.
This requires we have a mechanism that measures the distance between distributions.
The training strategy is illustrated through Fig.~\ref{fig:training scheme}, where the red lines indicate the training information flow.
Here we adopt similar structures to Wasserstein GAN (WGAN) \citep{Arjovsky17} to measure the Wasserstein distance between real samples and approximations.
Utilizing the Kantorovich-Rubinstein duality the Wasserstein distance bewteen $(x_t)$ and $(\hat{x}_t)$ can be written as 
\[\max_{D\in 1-Lip}\left(\mbbE[\mathbf{D}(\mathbf{x}^{(n)}_{t})]-\mbbE[\mathbf{D}(\hat{x}_t)]\right),\]
where $D$ is the discriminator function that is optimized under certain constraint.
Following the same line, we use two discriminators $D^{(n)}_\gamma,D^{(n)}_\omega$ to measure the Wasserstein distance between estimated weak innovations $(\nu_t)$ and i.i.d uniform sequence $(u_t)$ and between the original time series $(x_t)$ and the reconstruction $(\hat{x}_t)$, respectively.
Thus, by the fact that Wasserstein distance metricizes weak convergence, we design the following loss to obtain weak innovations auto-encoder:
\begin{multline}
     L_m^{(n)}((x_t),\theta_m,\eta_m):= \max_{\gamma_m,\omega_m}\big(\mbbE[\mathbf{D}^{(n)}_\gamma(\mathbf{u}^{(n)}_t) - \mbbE[\mathbf{D}^{(n)}_\gamma(\mathbf{G}^{(n)}_{\theta_m}(\mathbf{x}^{(n)}_t))]] \\+ \lambda(\mbbE[\mathbf{D}^{(n)}_{\omega_m}(\mathbf{x}^{(n)}_{t})]-\mbbE[\mathbf{D}^{(n)}_{\omega_m}(\mathbf{H}_\eta(\mathbf{G}^{(n)}_\theta(\mathbf{x}^{(n)}_{t,m})))])\big).
  \label{eq:loss}
\end{multline}

The two parts of the loss function regularize the innovations representation according to Eq.~\eqref{eq:s-iid} and Eq.~\eqref{Eq:recons}.
Therefore, by minimizing Eq.~\eqref{eq:loss} with respect to $\theta_m$ and $\eta_m$, one should be able to obtain the innovations sequence.

The finite dimensional training of WIAE does not guarantee to produce i.i.d. uniform innovations even with infinite amount of training data and the optimization in Eq.~\eqref{eq:loss} achieves global optimum.   Here we establish a structural convergence results that, as the dimension of the autoencoder increases, the sequence of finite dimensional WIAEs convergences to $(G,H)$ in the sense that the the sequence of weak innovations representation converges to the actual representation.  To make this precise, we introduce the following definition.

Let the superscipt $*$ denote the optimal quantities obtained via minimizing Eq.~\eqref{eq:loss}, $\forall m,n\in\mathbb{Z}^+$. 
Though we would like to have the entire sequence $(\nu_{t,m})\stackrel{d}{\rightarrow}(u_t)$ and $(\hat{x}_{t,m})\stackrel{d}{\rightarrow}(x_t)$, this cannot be enforced through finite-dimensional training objective.
Therefore, we define the convergence in distribution of finite block, which indicates the convergence in joint distribution of a finite dimension random vector.
\begin{definition}[Convergence in distribution of finite block $n$]
For all $n\in\mathbb{Z}^+$ fixed, an $m$-dimensional WIAE $\mathcal{A}^{(n)}_m$ trained with $n$-dimensional discriminators converges in distribution of finite block $n$ to $\mathcal{A} = (G, H)$ if, for all $t$,
\begin{align}
    \nubf^{(n)*}_{t,m} \stackrel{d}{\rightarrow} \nubf^{(n)}_t, \hat{\boldsymbol{x}}^{(n)*}_{t,m} \stackrel{d}{\rightarrow} \boldsymbol{x}^{(n)}_t,
\end{align}
as $m\rightarrow\infty$.
\label{def:convergence}
\end{definition}
Convergence in finite-training block is a compromised version of convergence in distribution for infinite series catered to finite dimension implementation of the discriminators.
In reality, the training block size $n$ can be chosen with respect to specific requirements and temporal dependency characteristics of certain application.

To achieve convergence of training block $n$ for some $n$ fixed, we make the following assumptions:
\ben
\item[A1] {\bf Existence:}  The random process $(x_t)$  has a weak innovations representation defined in (\ref{Eq:encoder} - \ref{Eq:decoder}), and there exists a causal encoder-decoder pair $(G, H)$ satisfying (\ref{Eq:encoder} - \ref{Eq:decoder}) with $H$ continuous.
\item[A2] {\bf Feasibility:} There exists a sequence of finite-dimensional auto-encoder functions $(G_{\tilde{\theta}_m}, H_{\tilde{\eta}_m})$ that converges uniformly to $(G,H)$ as $m\rightarrow \infty.$
\item[A3] {\bf Training:} The training sample sizes are  infinite. The training algorithm for all finite dimensional WIAE using finite dimensional training samples converges almost surely to the global optimal.
\een

\begin{theorem} 
\label{thm:converge}
Under (A1-A3),  $\Ac_{m}^{(n)}$ converges (of finite block size $n$) to $\Ac$, $\forall n \in \mathbb{Z}+.$
\end{theorem}
The proof is in the supplementary material.

\subsection{Probabilistic Forecasting and Sufficiency}
\label{subsec:inference}
Fig.~\ref{fig:prediction} illustrates the strategy of a $s$-step prediction, once the WIAE is trained.
Given past observations $\Xmsc_{t^*} := \{x_1,\cdots,x_{t^*}\}$, we can use the causal encoder $G_{\theta_m^*}$ to compute the weak innovations representations $(\nu_{1},\cdots,\nu_{t^*})$ up to time $t^*$.
To make $T$-step forecasts through the trained decoder $H_{\eta_m^*}$, the "new" information contained in unrealized innovations representation $\{\nu_{t^*+1},\cdots,\nu_{t^*+T}\}$ is missing.
By the definition of weak innovations representation, we know that the missing weak innovations are of i.i.d uniform distribution.
Thus, the conditional distribution of $x_{t^*}$ conditioned on $\Xmsc_{t^*-1}$ can be empirically obtained by repeatedly sampling the weak innovation of future timestamps $(\Tilde{\nu}_{t^*+1:t^*+T})$ from i.i.d uniform distribution.
For each realization of $(\Tilde{\nu}_{t^*+1:t^*+T})$, we can reconstruct one point of $\hat{x}_{t^*+T}$ that has the same distribution as $x_{t^*+T}$.
Through Monte-Carlo sampling of $\{\Tilde{\nu}_{T-s},\Tilde{\nu}_{T-s-1},\cdots\}$, one can reconstruct the distribution of $\hat{x}_{t}$ empirically, and consequently compute any statistics of interest.

We claim that learning weak innovations representation is sufficient for time series forecasting applications, which means that the predictions based on the weak innovations representation is equally well as the predictions made based on the original time series.
We verify the claim by showing that the conditional distribution of $x_t$ conditioned on $\Xmsc_{t-1}$ equals almost everywhere to the condition distribution $\hat{x}_t$ conditioned on $(\nu_s)_{s<t}$ under a mild regularity condition on $H$.
For the ease of notation, we denote the conditional cumulative distribution function of $(x_t)$ by
\[F(x|a_{t-1},a_{t-2},\cdots)=\mathbb{P}[x_t\leq x|x_{t-1}=a_{t-1},x_{t-2}=a_{t-2},\cdots],\]
and the counterpart of $\{\hat{x}_t\}$ by
\[\hat{F}(x|b_{t-1},b_{t-2},\cdots)=\mathbb{P}[\hat{x}_t\leq x|\nu_{t-1}=b_{t-1},\nu_{t-2}=b_{t-2},\cdots].\]

\paragraph{Assumption.} Let $(x_t)$ be a stationary time series for which the weak innovations (as defined in  Eq.~(\ref{Eq:encoder}-\ref{Eq:recons})) exists. Further, assume that the decoder function $H$ is injective.

\begin{theorem}[Sufficiency of Weak Innovations]
Let $(\nu_t)$ be the weak innovations representation of $(x_t)$ as defined in Eq.~(\ref{Eq:encoder}-\ref{Eq:recons}), satisfying the assumptions. Then, the conditional distribution function $F$ and $\hat{F}$, as a function of $x$, equals almost everywhere for almost all $a_{n-1},a_{n-2},\cdots$ and $b_{s} = H^{-1}(a_s,a_{s-1},\cdots)$.
In other word,
\[F(x|a_{t-1},a_{t-2},\cdots)=\hat{F}(x|b_{t-1},b_{t-2},\cdots) \] almost everywhere for almost all $a_{n-1},a_{n-2},\cdots$ and $b_{s} = H^{-1}(a_s,a_{s-1},\cdots)$.
\label{thm:sufficiency}
\end{theorem}

The proof of the theorem can be found in the appendix.
The significance of the theorem is that it allows us to conclude that for $(x_t)$ that satisfies the assumption, optimally trained WIAE is able to reconstruct the conditional distribution through Monte-Carlo sampling of the weak innovations representation.

\section{Simulation}
\label{sec:simulation}
\subsection{Representation Learning Evaluation}
\label{subsec:rep-learning}
We first evaluate the performance in terms of extracting weak innovations representation from original samples. 
Here we adopted the hypothesis testing formulation known as the Runs Up and Down test \citep{Gibbons:03Book}. 
Its null hypothesis assumes the sequence being i.i.d, and the alternative hypothesis the opposite. 
The test was based on collecting the number of consecutively increasing or decreasing subsequences, and then calculate the p-value of the test based on its asymptotic distribution.  According to \citet{Gibbons:03Book}, the runs up and down test had empirically the best performance. 

We compared with $5$ representation learning benchmarks: fAnoGAN \citep{Schlegl&Seebock:19}, Anica \citep{Brakel&Bengio:17}, IAE \citep{WangTong:21JMLR}, Pyraformer \citep{liu_pyraformer_2022}, TLAE \citep{nguyen_temporal_2021}, on both synthetic datasets (LAR, MA, MC) and real-world datsets (ISONE, NYISO, SP500).
LAR, MA, MC stands for Linear Autoregression, Moving Average and Markov Chain models, respectively.
For the LAR case, \cite{Wu:05PNAS} showed that the original form of innovations (Eq.~(\ref{eq:s-encoder}-\ref{eq:s-iid})) exist, while for the MA case that is non-minimum phase, the existence remains unknown.
For the MC case, \cite{Rosenblatt:59} proved that the original form of the innovations representation doesn't exist, though the weak innovations do.
More details of benchmarks and datasets can be found in the appendix.
\renewcommand{\arraystretch}{0.5}
\begin{table}[t]
\fontsize{6}{0.1}
    \centering
    \caption{P-value from Runs Test, Wasserstein distance (WassDist-rep) bewteen representations and uniform {\it i.i.d} sequence, and Wasserstein distance between the original sequence and reconstruction (WassDist-recons).}
    \begin{tabular}{ccccccc}
    \toprule
       {\scriptsize Metrics} & {\scriptsize Datasets} & {\scriptsize WIAE} & \scriptsize IAE &\scriptsize fAnoGAN &\scriptsize Anica &\scriptsize TLAE  \\
        \midrule
         \multirow{6}{*}{\scriptsize P-value} 
         &\scriptsize LAR   &$\mathbf{0.9513}$ &$0.8837$ &$0.0176$   &$<0.001$   &$0.4902$\\
         &\scriptsize MA    &$0.8338$ &$\mathbf{0.9492}$ &$0.2248$   &$<0.001$   &$0.5498$\\
         & \scriptsize MC    &$\mathbf{0.7651}$ &$<0.001$ &$0.2080$   &$<0.001$   &$<0.001$\\
         &\scriptsize ISONE &$\mathbf{0.8897}$ &$<0.001$ &$<0.001$   &$0.2610$   &$0.6731$\\
         & \scriptsize NYISO &$\mathbf{0.7943}$ &$<0.001$ &$<0.001$   &$<0.001$   &$0.2269$\\
         &\scriptsize SP500 &$\mathbf{0.8764}$ &$<0.001$ &$0.4977$   &$0.2655$   &$0.8183$\\
         \midrule
         \multirow{6}{*}{\shortstack{\scriptsize WassDist-rep\\$\pm$ \scriptsize std}} 
         &\scriptsize LAR   &$\mathbf{0.3651\pm0.0033}$  &$0.5555\pm0.0551$  &$0.9289\pm0.0207$  &$0.8698\pm0.0186$  &$0.9525\pm0.1697$\\
         &\scriptsize MA    &$\mathbf{0.3563\pm0.0053}$  &$0.4839\pm0.0202$  &$1.0679\pm0.0339$  &$0.8518\pm0.0740$  &$0.7346\pm0.2502$\\
         &\scriptsize MC    &$0.3814\pm0.0053$  &$0.3984\pm0.0687$  &$1.3093\pm0.0276$  &$0.7112\pm0.0231$  &$\mathbf{0.0906\pm0.6693}$\\
         &\scriptsize ISONE &$\mathbf{0.0391\pm0.0243}$  &$0.3796\pm0.2080$  &$0.0959\pm0.0980$  &$1.0579\pm0.1328$  &$0.0814\pm0.0745$\\
         &\scriptsize NYISO &$\mathbf{0.0405\pm0.0306}$  &$0.3192\pm0.1186$  &$0.1102\pm0.0322$  &$1.4802\pm0.1069$  &$0.0678\pm0.0239$\\
         &\scriptsize SP500 &$\mathbf{0.1287\pm0.0036}$  &$0.3778\pm0.1904$  &$0.1341\pm0.0538$  &$1.1658\pm0.1398$  &$0.1755\pm0.1478$\\
         \midrule
         \multirow{6}{*}{\shortstack{\scriptsize WassDist-recons\\$\pm$ \scriptsize std}} 
         &\scriptsize LAR   &$1.7399\pm0.1464$  &$\mathbf{1.7119\pm0.1110}$  &$2.5150\pm1.0666$  &$4.0758\pm1.1990$  &$1.7625\pm0.1280$\\
         &\scriptsize MA    &$\mathbf{2.0214\pm0.1063}$  &$2.6697\pm0.5221$  &$10.9907\pm0.2981$ &$19.3772\pm0.8062$ &$4.6101\pm0.1878$\\
         &\scriptsize MC    &$\mathbf{0.1960\pm0.0338}$  &$2.8264\pm0.0996$  &$11.5251\pm1.7173$ &$12.4828\pm1.5741$ &$0.2639\pm0.1545$\\
         &\scriptsize ISONE &$\mathbf{0.3382\pm0.0695}$  &$0.8505\pm0.0032$  &$1.1341\pm0.05174$ &$2.0250\pm0.1790$  &$0.7436\pm0.1592$\\
         &\scriptsize NYISO &$0.1217\pm0.0029$  &$\mathbf{0.1111\pm0.0070}$  &$1.4021\pm0.07022$ &$2.3860\pm0.1307$  &$1.4098\pm0.3289$\\
         &\scriptsize SP500 &$\mathbf{0.1677\pm0.0551}$  &$1.1021\pm0.1904$  &$0.8563\pm0.2455$  &$1.8874\pm0.3518$  &$1.6977\pm0.4224$\\
         \bottomrule
    \end{tabular}
    \label{tab:property}
\end{table}

The results of property tests are shown in Table.~\ref{tab:property}. 
Seen from the results, only WIAE had the capability of consistently producing i.i.d sequences that couldn't be rejected by the runs test for all cases, with small Wasserstein distances between representation and uniform i.i.d  sequence, and between the original series and reconstruction.
For all other techniques, at lease one case exists for which either the sequences they produced can be easily rejected by the runs test, or the sequences had high Wasserstein distances.
This indicated that the WIAE architecture design is necessary for the extraction of weak innovations representation, and that currently no such technique that fulfills the requirement apart from WIAE.
 


\subsection{Probabilistic Forecasting}
\label{subsec:forecast}
\renewcommand{\arraystretch}{0.4}
\begin{table}[t]
\fontsize{7.5}{1}
  \caption{Numerical Results. The numbers in the parenthese indicate the prediction step.}
  \label{tb:Result-1}
  \centering
  \begin{tabular}{lllllllllll}
    \toprule
    \scriptsize Metrics &\scriptsize Method     &\scriptsize ISONE($15$)    &\scriptsize ISONE($24$) &\scriptsize NYISO($15$) &\scriptsize NYISO($24$) &\scriptsize SP500($2$) \\
    \midrule
       \scriptsize \multirow{6}{*}{\textbf{NMSE}} & \scriptsize WIAE &$\mathbf{0.0857}$  & $\mathbf{0.0868}$ &$\mathbf{0.0876}$ &$\mathbf{0.08886}$  &$0.0049$   \\
    &\scriptsize DeepAR    &$0.1064$ &$0.2852$ &$0.0952$ &$0.1632$ &$\mathbf{0.0002}$    \\
    &\scriptsize NPTS     &$0.1216$ &$0.1250$ &$0.1840$ &$0.2031$ & $0.0003$   \\
    &\scriptsize Pyraformer &$0.1247$ &$0.1779$ &$0.4031$ &$0.4229$ &$0.0050$\\
    &\scriptsize TLAE &$0.1219$ &$0.1837$ &$0.1116$ &$0.1492$ &$0.0013$\\
    &\scriptsize Wavenet &$0.1245$ &$0.1643$ &$0.1128$ &$0.1809$ &$0.0236$\\
    &\scriptsize SNARX &$0.8632$ &$0.9658$ &$0.9586$ &$1.8063$ &$0.4193$\\
    \midrule
    \scriptsize \multirow{6}{*}{\textbf{NMeSE}}
    &\scriptsize WIAE &$0.0238$  & $\mathbf{0.0257}$ &$0.0284$ &$0.0283$  &$0.0017$   \\
    &\scriptsize DeepAR    &$0.0452$ &$0.1187$ &$\mathbf{0.0103}$ &$\mathbf{0.0164}$ &$0.0001$    \\
    &\scriptsize NPTS     &$0.0493$ &$0.0567$ &$0.0250$ &$0.0282$ & $\mathbf{0.00003}$   \\
    &\scriptsize Pyraformer &$0.0266$ &$0.0285$ &$0.1053$ &$0.1633$ &$0.0047$\\
    &\scriptsize TLAE &$\mathbf{0.0148}$ &$0.0293$ &$0.0190$ &$0.0260$ &$0.0007$\\
    &\scriptsize Wavenet &$0.0469$ &$0.0661$ &$0.0227$ &$0.0293$ &$0.0003$\\
    &\scriptsize SNARX &$0.1053$ &$0.0996$ &$0.0296$ &$0.3486$ &$0.2980$\\
    \midrule
    \scriptsize \multirow{6}{*}{\textbf{NMAE}}
    &\scriptsize WIAE &$\mathbf{0.2327}$  & $\mathbf{0.2330}$ &$0.2112$ &$\mathbf{0.2104}$  &$0.0538$   \\
    &\scriptsize DeepAR    &$0.2969$ &$0.4458$ &$0.1799$ &$0.2451$ &$0.0108$    \\
    &\scriptsize NPTS     &$0.2896$ &$0.2958$ &$0.2260$ &$0.2138$ & $\mathbf{0.0100}$   \\
    &\scriptsize Pyraformer &$0.2570$ &$0.3518$ &$0.4892$ &$0.5234$ &$0.4819$\\
    &\scriptsize TLAE  &$0.2570$ &$0.3536$ &$0.2258$ &$0.2590$ &$0.0313$\\
    &\scriptsize Wavenet  &$0.3154$ &$0.3310$ &$\mathbf{0.1770}$ &$0.2330$ &$0.0552$\\
    &\scriptsize SNARX  &$0.8317$ &$0.9611$ &$0.9299$ &$0.9967$ &$0.6135$\\
    \midrule
    \scriptsize \multirow{6}{*}{\textbf{NMeAE}}
    &\scriptsize WIAE &$0.1693$  & $\mathbf{0.1653}$ &$0.1635$ &$0.1643$  &$0.0417$   \\
    &\scriptsize DeepAR    &$0.2271$ &$0.3786$ &$\mathbf{0.0991}$ &$0.1268$ &$0.0085$    \\
    &\scriptsize NPTS     &$0.2146$ &$0.2299$ &$0.1243$ &$\mathbf{0.1256}$ & $\mathbf{0.0056}$   \\
    &\scriptsize Pyraformer &$0.1815$ &$0.1953$ &$0.3320$ &$0.3671$ &$0.1926$\\
    &\scriptsize TLAE  &$\mathbf{0.1135}$ &$0.2292$ &$0.1164$ &$0.1722$ &$0.0273$\\
    &\scriptsize Wavenet &$0.2384$ &$0.2580$ &$0.1401$ &$0.0552$ &$0.0178$\\
    &\scriptsize SNARX   &$0.5105$ &$0.4981$ &$0.2893$ &$0.9962$ &$0.5793$\\
    \midrule
    \scriptsize \multirow{6}{*}{\textbf{MASE}}
    &\scriptsize WIAE &$\mathbf{1.0442}$  &$\mathbf{0.8012}$ &$0.9526$ &$\mathbf{0.8468}$  &$6.4547$   \\
    &\scriptsize DeepAR    &$1.6767$ &$2.5091$ &$\mathbf{0.8383}$ &$1.2080$ &$0.0085$    \\
    &\scriptsize NPTS     &$1.3199$ &$1.4103$ &$0.1792$ &$1.2807$ & $\mathbf{1.1475}$   \\
    &\scriptsize Pyraformer  &$1.2277$ &$1.1337$ &$1.6397$ &$1.7081$ &$17.4409$\\
    &\scriptsize TLAE   &$1.2130$ &$1.1190$ &$1.0656$ &$0.9851$ &$1.4090$ \\
    &\scriptsize Wavenet &$1.3912$ &$1.1275$ &$0.9966$ &$1.4527$ &$6.2926$\\
    &\scriptsize SNARX   &$1.7154$   &$1.2756$   &$1.1210$    &$2.8268$ &$21.9792$\\
    \midrule
    \scriptsize \multirow{6}{*}{\textbf{sMAPE}}
    &\scriptsize WIAE &$0.2421$  & $\mathbf{0.2419}$ &$0.2106$ &$\mathbf{0.2091}$  &$0.0540$   \\
    &\scriptsize DeepAR    &$\mathbf{0.2372}$ &$0.7076$ &$\mathbf{0.1709}$ &$0.2236$ &$\mathbf{0.0106}$    \\
    &\scriptsize NPTS     &$0.8601$ &$0.8854$ &$0.7395$ &$0.7506$ & $0.0518$   \\
    &\scriptsize Pyraformer  &$0.2785$ &$0.2896$ &$0.3672$ &$0.4095$ &$0.2785$\\
    &\scriptsize TLAE   &$0.2440$ &$0.3137$ &$0.1974$ &$0.2549$ &$0.0298$ \\
    &\scriptsize Wavenet &$0.3057$ &$0.3375$ &$0.2090$ &$0.2475$ &$0.0480$\\
    &\scriptsize SNARX  &$0.5035$ &$1.3589$ &$1.2661$ &$1.8736$ &$0.8952$\\
    \bottomrule
  \end{tabular}
\end{table}
We compare our methods with $6$ other state-of-art time series forecasting techniques: DeepAR \citep{salinas_deepar_2019}, Non-Parametric Time Series Forecaster (NPTS) \citep{alexandrov_gluonts_2019}, Pyraformer \citep{liu_pyraformer_2022}, TLAE \citep{nguyen_temporal_2021}, Wavenet \citep{oord_wavenet_2016}, and SNARX \citep{weron_forecasting_2008}.
DeepAR is an auto-regressive RNN time series model that estimates parameters of parametric distributions.
NPTS is a probabilistic forecasting technique that resembles naive forecasters.
It randomly samples a past time index following a categorical probability distribution over time indices.
TLAE is a non-parametric probabilistic forecasting technique that utilizes autoencoder architecture to learn the underlying Gaussian latent process, and uses it as the estimator.
Wavenet is a parametric probabilistic forecasting technique that is based on the dilated causal convolutions. 
Pyraformer is a point estimation technique that adopts multi-resolution attention modules, and is trained by minimizing mean squared error.
SNARX is a semi-parametric AR model that utilizes kernel density function to estimate the distribution of noise, which has superior emipirical performance on electricity pricing datasets according to \cite{weron_electricity_2014}.
\begin{figure}
    \centering
        \includegraphics[scale=0.4]{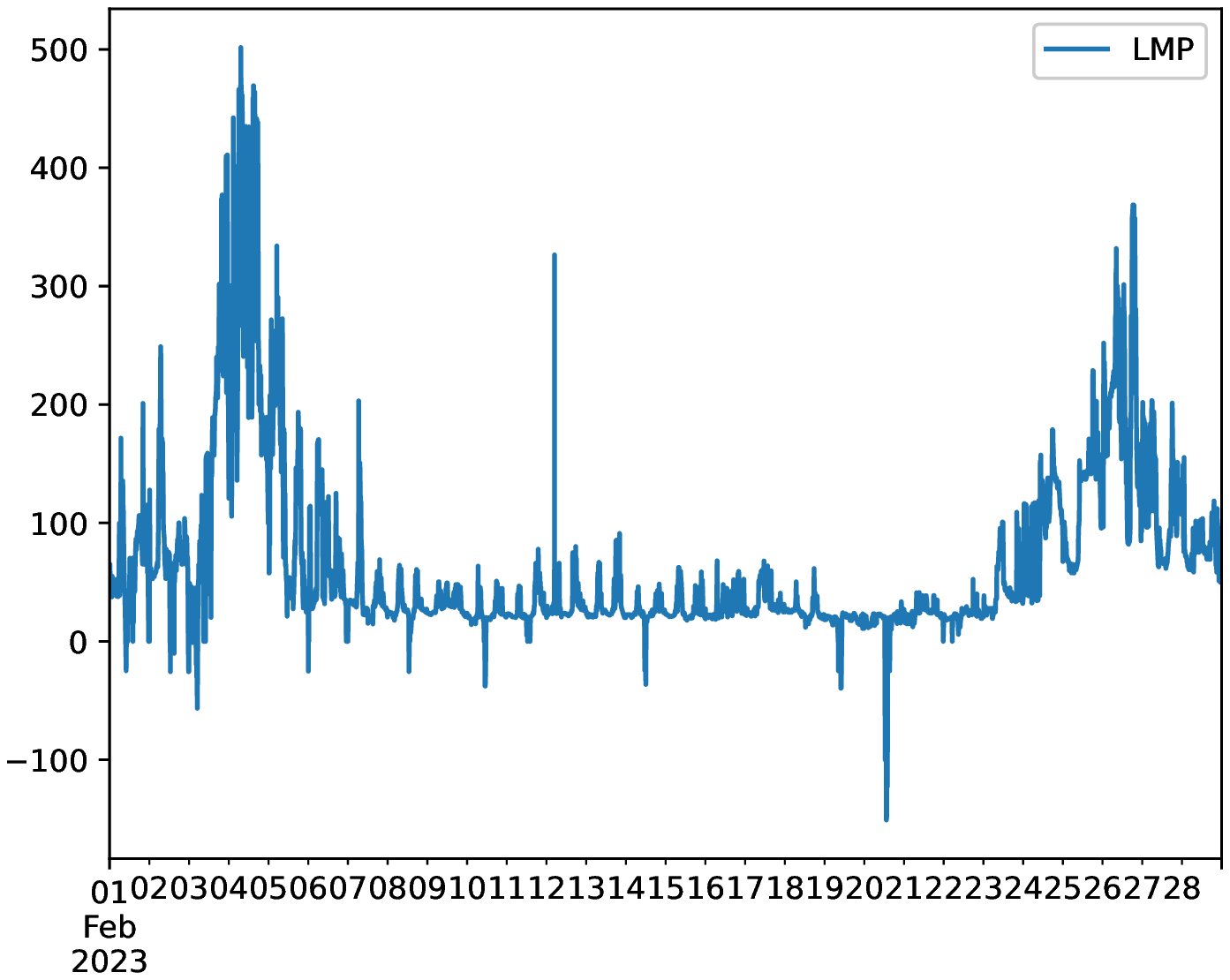}
        \includegraphics[scale=0.45]{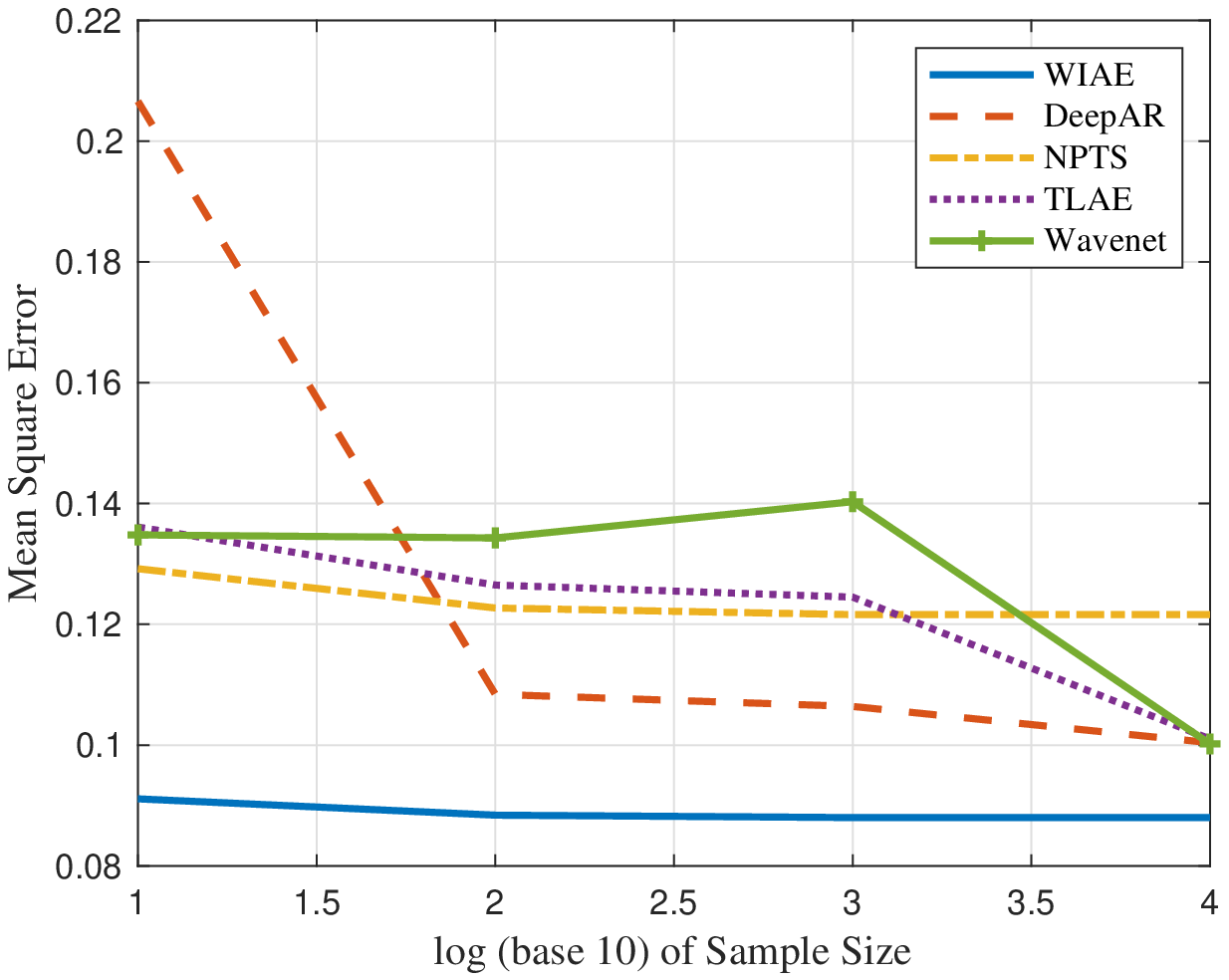}
    \caption{Trajectory of ISONE (right) and sensitivity analysis with respect to the number of repeatedly sampled trajectories.}
    \label{fig:traj}
\end{figure}
We simulated all methods on three commonly used datasets: {\it SP500}, {\it traffic} and {\it electricity}.
{\it Electricity} dataset consists of $15$-minute electricity consumption from $370$ households; {\it electricity} consists of hourly traffic sampled from $963$ car lanes; {\it SP500} contains daily opening and closing prices of SP500 stocks.
To demonstrate the ability of forecasting time series with higher sampling frequency and larger variability, we also choose three publicly available real-time electricity price from three independent system operators (ISO).
We named the datasets with the name of the ISOs: New York ISO (NYISO), ISO New England (ISONE) and PJM\footnote{ISO data can be found on the official webset.}.
NYISO and ISONE consists of 5-minute real time electricity prices calculated through optimization programs involving load demand, generators' bids and conditions, and network conditions in real time.
Due to changes in network conditions, real-time electricity prices are subject to rapid fluctuations, as demonstrated by Fig.~\ref{fig:traj} (left).
PJM dataset includes hourly electricity price that is calculated similarly. We took one month (February of 2023) of electricity prices for NYISO and ISONE, and one year (2022/01/01 - 2022/12/31) for PJM.

To fully demonstrate the advantage of probabilistic forecasting, we adopted multiple commonly used metrics: normalized mean square error (NMSE), normalized median square error (NmeSE), normalized mean absolute error (NMAE), normalized median absolute error (NmeAE), mean absolute squared error (MASE) and symmetric mean absolute percentage error (sMAPE). 
For probabilistic forecasting methods, we use sample mean as estimator when calculating square errors, and sample median for absolute errors.
$1000$ trajectories were sampled for probabilistic methods.
The commonly used mean absolute percentage error was not adopted, since in electricity dataset the absolute value of the actual value can be very close to $0$, which nullifies the effectiveness of the metrics.
Since electricity prices datasets exhibit high variability, we excluded the outliers, which are defined as samples that are three standard deviations away from sample mean, when calculating all metrics, for all methods.
Notice that for normalized absolute errors and normalized squared errors, we computed both their mean and median, due to the fact that median errors are more robust to outliers.
A comprehensive view of errors distribution can be obtained from their empirical cumulative distribution functions (eCDF), which is included in the supplementary material.

The simulation result is shown in Table.~\ref{tb:Result-1}.
It can be seen that WIAE had better performance for most of the metrics. 
Pyraformer, as a point estimation technique optimized by MSE, is worse under metrics using absolute errors.
We also observed that the auto-regressive probabilistic forecasting methods (DeepAR, NPTS, Wavenet,SNARX) have the tendency of being affected by past observation, which leads to better performance when the time series is smooth with few fluctuation.
However, for electricity price datasets where the variability is high, those auto-regressive methods are inclined to have larger errors.

\subsection{Sensitivity Analysis w.r.t Number of Repeated Sampling}
Most probabilistic forecasting techniques rely on repeated sampling of random variables to get a empirical joint distribution represented by traces.
The natural question to ask is that what is the minimum number of repeated sampling required to accurately represent the joint distribution.
In this section, we analyze the variation of probabilistic forecasting methods' performance with the number of repeatedly sampled trajectories trajectories.
NMSE was chosen to represent the level of performance, and $1000$ independent experiments were conducted to calculate NMSE for each scenario.
The results are shown in Fig.~\ref{fig:traj} (right).
It can be seen that WIAE reaches a higher accuracy at the sample size $100$, lower than most other methods.
This result demonstrated the benefit of WIAE that comes from using independence of weak innovations representation, for the number of trajectories needed to comprehensively represent a high-dimensional distribution with independent coordinates is significantly lower than that of distribution with dependent coordinates.

\section{Conclusion}
We presented a deep-learning solution to acquire the weak innovations representation, which is causally computed with temporal independence.
Our method's capability of extracting weak innovations representation was demonstrated both by the structural convergence theorem, and the numerical results.
Additionally, we made methodological advances by adopting the weak innovations representation for non-parametric time series forecasting problems.
Our method has been shown to work accurately as a time series forecasting technique, with better numerical performance on both smooth datasets and datasets with larger variability.
The independent temporal structure of weak innovations representation also benefits our method by reducing the amount of repeated sampling of trajectories to obtain accurate estimators, comparing with other probabilistic forecasting techniques.
Future directions include multivariate extension of our methods.

\newpage
\bibliography{ts_forecast}

\begin{thebibliography}{46}
\providecommand{\natexlab}[1]{#1}
\providecommand{\url}[1]{\texttt{#1}}
\expandafter\ifx\csname urlstyle\endcsname\relax
  \providecommand{\doi}[1]{doi: #1}\else
  \providecommand{\doi}{doi: \begingroup \urlstyle{rm}\Url}\fi

\bibitem[Alexandrov et~al.(2019)Alexandrov, Benidis, Bohlke-Schneider,
  Flunkert, Gasthaus, Januschowski, Maddix, Rangapuram, Salinas, Schulz,
  Stella, Türkmen, and Wang]{alexandrov_gluonts_2019}
A.~Alexandrov, K.~Benidis, M.~Bohlke-Schneider, V.~Flunkert, J.~Gasthaus,
  T.~Januschowski, D.~C. Maddix, S.~Rangapuram, D.~Salinas, J.~Schulz,
  L.~Stella, A.~C. Türkmen, and Y.~Wang.
\newblock {GluonTS}: {Probabilistic} {Time} {Series} {Models} in {Python}, June
  2019.
\newblock URL \url{http://arxiv.org/abs/1906.05264}.
\newblock arXiv:1906.05264 [cs, stat].

\bibitem[Arjovsky et~al.(2017)Arjovsky, Chintala, and L.Bottou]{Arjovsky17}
M.~Arjovsky, S.~Chintala, and L.Bottou.
\newblock {Wasserstein GAN}, Jan. 2017.
\newblock arXiv:1701.07875.

\bibitem[Auestad and Tjøstheim(1990)]{auestad_identification_1990}
B.~Auestad and D.~Tjøstheim.
\newblock Identification of {Nonlinear} {Time} {Series}: {First} {Order}
  {Characterization} and {Order} {Determination}.
\newblock \emph{Biometrika}, 77\penalty0 (4):\penalty0 669--687, 1990.
\newblock ISSN 0006-3444.
\newblock \doi{10.2307/2337091}.
\newblock URL \url{https://www.jstor.org/stable/2337091}.
\newblock Publisher: [Oxford University Press, Biometrika Trust].

\bibitem[Bertsekas(2012)]{bertsekas_dynamic_2012}
D.~P. Bertsekas.
\newblock \emph{Dynamic programming and optimal control}.
\newblock Athena Scientific, Belmont, Mass., fourth edition. edition, 2012.
\newblock ISBN 978-1-886529-44-1.
\newblock http://catalog.library.cornell.edu/catalog/9826722.

\bibitem[Borovykh et~al.(2018)Borovykh, Bohte, and
  Oosterlee]{borovykh_conditional_2018}
A.~Borovykh, S.~Bohte, and C.~W. Oosterlee.
\newblock Conditional {Time} {Series} {Forecasting} with {Convolutional}
  {Neural} {Networks}, Sept. 2018.
\newblock URL \url{http://arxiv.org/abs/1703.04691}.
\newblock arXiv:1703.04691 [stat].

\bibitem[Brakel and Bengio(2017)]{Brakel&Bengio:17}
P.~Brakel and Y.~Bengio.
\newblock {Learning Independent Features with Adversarial Nets for Non-linear
  ICA}, Oct. 2017.
\newblock arXiv:1710.05050.

\bibitem[Chan and Tong(1986)]{chan_estimating_1986}
K.~S. Chan and H.~Tong.
\newblock On {Estimating} {Thresholds} in {Autoregressive} {Models}.
\newblock \emph{Journal of Time Series Analysis}, 7\penalty0 (3):\penalty0
  179--190, 1986.
\newblock ISSN 1467-9892.
\newblock \doi{10.1111/j.1467-9892.1986.tb00501.x}.
\newblock URL
  \url{https://onlinelibrary.wiley.com/doi/abs/10.1111/j.1467-9892.1986.tb00501.x}.
\newblock \_eprint:
  https://onlinelibrary.wiley.com/doi/pdf/10.1111/j.1467-9892.1986.tb00501.x.

\bibitem[Du et~al.(2022)Du, Du, and Li]{du_probabilistic_2022}
H.~Du, S.~Du, and W.~Li.
\newblock Probabilistic time series forecasting with deep non-linear state
  space models.
\newblock \emph{CAAI Transactions on Intelligence Technology}, n/a\penalty0
  (n/a), 2022.
\newblock ISSN 2468-2322.
\newblock \doi{10.1049/cit2.12085}.
\newblock URL \url{https://onlinelibrary.wiley.com/doi/abs/10.1049/cit2.12085}.
\newblock \_eprint:
  https://ietresearch.onlinelibrary.wiley.com/doi/pdf/10.1049/cit2.12085.

\bibitem[Gibbons and Chakraborti(2003)]{Gibbons:03Book}
J.~D. Gibbons and S.~Chakraborti.
\newblock \emph{{Nonparametric Statistical Inference. (4th ed., rev. and
  expanded.)}}.
\newblock M. Dekker, New York, 2003.

\bibitem[Granger and Teräsvirta(1993)]{granger_modelling_1993}
C.~W.~J. Granger and T.~Teräsvirta.
\newblock \emph{Modelling {Nonlinear} {Economic} {Relationships}}.
\newblock Advanced {Texts} in {Econometrics}. Oxford University Press, Oxford,
  New York, Dec. 1993.
\newblock ISBN 978-0-19-877320-7.

\bibitem[Haggan and Ozaki(1981)]{haggan_modelling_1981}
V.~Haggan and T.~Ozaki.
\newblock Modelling {Nonlinear} {Random} {Vibrations} {Using} an
  {Amplitude}-{Dependent} {Autoregressive} {Time} {Series} {Model}.
\newblock \emph{Biometrika}, 68\penalty0 (1):\penalty0 189--196, 1981.
\newblock ISSN 0006-3444.
\newblock \doi{10.2307/2335819}.
\newblock URL \url{https://www.jstor.org/stable/2335819}.
\newblock Publisher: [Oxford University Press, Biometrika Trust].

\bibitem[Hart(1996)]{hart_automated_1996}
J.~D. Hart.
\newblock Some automated methods of smoothing time-dependent data.
\newblock \emph{Journal of Nonparametric Statistics}, 6\penalty0
  (2-3):\penalty0 115--142, Jan. 1996.
\newblock ISSN 1048-5252.
\newblock \doi{10.1080/10485259608832667}.
\newblock URL \url{https://doi.org/10.1080/10485259608832667}.
\newblock Publisher: Taylor \& Francis \_eprint:
  https://doi.org/10.1080/10485259608832667.

\bibitem[Härdle and Vieu(1992)]{hardle_kernel_1992}
W.~Härdle and P.~Vieu.
\newblock Kernel {Regression} {Smoothing} of {Time} {Series}.
\newblock \emph{Journal of Time Series Analysis}, 13\penalty0 (3):\penalty0
  209--232, 1992.
\newblock ISSN 1467-9892.
\newblock \doi{10.1111/j.1467-9892.1992.tb00103.x}.
\newblock URL
  \url{https://onlinelibrary.wiley.com/doi/abs/10.1111/j.1467-9892.1992.tb00103.x}.
\newblock \_eprint:
  https://onlinelibrary.wiley.com/doi/pdf/10.1111/j.1467-9892.1992.tb00103.x.

\bibitem[Härdle et~al.(1997)Härdle, Lütkepohl, and Chen]{hardle_review_1997}
W.~Härdle, H.~Lütkepohl, and R.~Chen.
\newblock A {Review} of {Nonparametric} {Time} {Series} {Analysis}.
\newblock \emph{International Statistical Review / Revue Internationale de
  Statistique}, 65\penalty0 (1):\penalty0 49--72, 1997.
\newblock ISSN 0306-7734.
\newblock \doi{10.2307/1403432}.
\newblock URL \url{https://www.jstor.org/stable/1403432}.
\newblock Publisher: [Wiley, International Statistical Institute (ISI)].

\bibitem[Härdle et~al.(1998)Härdle, Tsybakov, and
  Yang]{hardle_nonparametric_1998}
W.~Härdle, A.~Tsybakov, and L.~Yang.
\newblock Nonparametric vector autoregression.
\newblock \emph{Journal of Statistical Planning and Inference}, 68\penalty0
  (2):\penalty0 221--245, May 1998.
\newblock ISSN 0378-3758.
\newblock \doi{10.1016/S0378-3758(97)00143-2}.
\newblock URL
  \url{https://www.sciencedirect.com/science/article/pii/S0378375897001432}.

\bibitem[{Kailath}(1968)]{Kailath1968TAC}
T.~{Kailath}.
\newblock {An Innovations Approach to Least-Squares Estimation--{P}art {I}:
  Linear Filtering in Additive White Noise}.
\newblock \emph{IEEE Transactions on Automatic Control}, 13\penalty0
  (6):\penalty0 646--655, 1968.
\newblock \doi{10.1109/TAC.1968.1099025}.

\bibitem[Kailath(2000)]{kailath_linear_2000}
T.~Kailath.
\newblock \emph{Linear estimation}.
\newblock Prentice Hall, Upper Saddle River, N.J., 2000.
\newblock ISBN 978-0-13-022464-4.
\newblock http://catalog.library.cornell.edu/catalog/4369884.

\bibitem[Kalman(1960)]{Kalman:60TASME}
R.~E. Kalman.
\newblock {A New Approach to Linear Filtering and Prediction Problems}.
\newblock \emph{{Trans. ASME J. of Basic Engineering}}, 82\penalty0
  (1):\penalty0 35--45, 1960.

\bibitem[LE~GUEN and THOME(2020)]{le_guen_probabilistic_2020}
V.~LE~GUEN and N.~THOME.
\newblock Probabilistic {Time} {Series} {Forecasting} with {Shape} and
  {Temporal} {Diversity}.
\newblock In \emph{Advances in {Neural} {Information} {Processing} {Systems}},
  volume~33, pages 4427--4440. Curran Associates, Inc., 2020.
\newblock URL
  \url{https://proceedings.neurips.cc/paper/2020/hash/2f2b265625d76a6704b08093c652fd79-Abstract.html}.

\bibitem[Li et~al.(2021)Li, Zhang, Yan, Jin, Zhang, Duan, and
  Tian]{li_synergetic_2021}
L.~Li, J.~Zhang, J.~Yan, Y.~Jin, Y.~Zhang, Y.~Duan, and G.~Tian.
\newblock Synergetic {Learning} of {Heterogeneous} {Temporal} {Sequences} for
  {Multi}-{Horizon} {Probabilistic} {Forecasting}.
\newblock \emph{Proceedings of the AAAI Conference on Artificial Intelligence},
  35\penalty0 (10):\penalty0 8420--8428, May 2021.
\newblock ISSN 2374-3468.
\newblock \doi{10.1609/aaai.v35i10.17023}.
\newblock URL \url{https://ojs.aaai.org/index.php/AAAI/article/view/17023}.
\newblock Number: 10.

\bibitem[Lim et~al.(2020)Lim, Arik, Loeff, and Pfister]{lim_temporal_2020}
B.~Lim, S.~O. Arik, N.~Loeff, and T.~Pfister.
\newblock Temporal {Fusion} {Transformers} for {Interpretable} {Multi}-horizon
  {Time} {Series} {Forecasting}, Sept. 2020.
\newblock URL \url{http://arxiv.org/abs/1912.09363}.
\newblock arXiv:1912.09363 [cs, stat].

\bibitem[Liu et~al.(2022)Liu, Yu, Liao, Li, Lin, Liu, and
  Dustdar]{liu_pyraformer_2022}
S.~Liu, H.~Yu, C.~Liao, J.~Li, W.~Lin, A.~X. Liu, and S.~Dustdar.
\newblock Pyraformer: {Low}-{Complexity} {Pyramidal} {Attention} for
  {Long}-{Range} {Time} {Series} {Modeling} and {Forecasting}.
\newblock Mar. 2022.
\newblock URL \url{https://openreview.net/forum?id=0EXmFzUn5I}.

\bibitem[Masry and Tjøstheim(1995)]{masry_nonparametric_1995}
E.~Masry and D.~Tjøstheim.
\newblock Nonparametric {Estimation} and {Identification} of {Nonlinear} {ARCH}
  {Time} {Series}: {Strong} {Convergence} and {Asymptotic} {Normality}.
\newblock \emph{Econometric Theory}, 11\penalty0 (2):\penalty0 258--289, 1995.
\newblock ISSN 0266-4666.
\newblock URL \url{https://www.jstor.org/stable/3532573}.
\newblock Publisher: Cambridge University Press.

\bibitem[Nguyen and Quanz(2021)]{nguyen_temporal_2021}
N.~Nguyen and B.~Quanz.
\newblock Temporal {Latent} {Auto}-{Encoder}: {A} {Method} for {Probabilistic}
  {Multivariate} {Time} {Series} {Forecasting}.
\newblock \emph{Proceedings of the AAAI Conference on Artificial Intelligence},
  35\penalty0 (10):\penalty0 9117--9125, May 2021.
\newblock ISSN 2374-3468.
\newblock \doi{10.1609/aaai.v35i10.17101}.
\newblock URL \url{https://ojs.aaai.org/index.php/AAAI/article/view/17101}.
\newblock Number: 10.

\bibitem[Nie et~al.(2023)Nie, Nguyen, Sinthong, and Kalagnanam]{nie_time_2023}
Y.~Nie, N.~H. Nguyen, P.~Sinthong, and J.~Kalagnanam.
\newblock A {Time} {Series} is {Worth} 64 {Words}: {Long}-term {Forecasting}
  with {Transformers}.
\newblock Feb. 2023.
\newblock URL \url{https://openreview.net/forum?id=Jbdc0vTOcol}.

\bibitem[Oord et~al.(2016)Oord, Dieleman, Zen, Simonyan, Vinyals, Graves,
  Kalchbrenner, Senior, and Kavukcuoglu]{oord_wavenet_2016}
A.~v.~d. Oord, S.~Dieleman, H.~Zen, K.~Simonyan, O.~Vinyals, A.~Graves,
  N.~Kalchbrenner, A.~Senior, and K.~Kavukcuoglu.
\newblock {WaveNet}: {A} {Generative} {Model} for {Raw} {Audio}, Sept. 2016.
\newblock URL \url{http://arxiv.org/abs/1609.03499}.
\newblock arXiv:1609.03499 [cs].

\bibitem[Rangapuram et~al.(2018)Rangapuram, Seeger, Gasthaus, Stella, Wang, and
  Januschowski]{rangapuram_deep_2018}
S.~S. Rangapuram, M.~W. Seeger, J.~Gasthaus, L.~Stella, Y.~Wang, and
  T.~Januschowski.
\newblock Deep {State} {Space} {Models} for {Time} {Series} {Forecasting}.
\newblock In \emph{Advances in {Neural} {Information} {Processing} {Systems}},
  volume~31. Curran Associates, Inc., 2018.
\newblock URL
  \url{https://papers.nips.cc/paper/2018/hash/5cf68969fb67aa6082363a6d4e6468e2-Abstract.html}.

\bibitem[Robinson(1983)]{robinson_nonparametric_1983}
P.~M. Robinson.
\newblock Nonparametric {Estimators} for {Time} {Series}.
\newblock \emph{Journal of Time Series Analysis}, 4\penalty0 (3):\penalty0
  185--207, 1983.
\newblock ISSN 1467-9892.
\newblock \doi{10.1111/j.1467-9892.1983.tb00368.x}.
\newblock URL
  \url{https://onlinelibrary.wiley.com/doi/abs/10.1111/j.1467-9892.1983.tb00368.x}.
\newblock \_eprint:
  https://onlinelibrary.wiley.com/doi/pdf/10.1111/j.1467-9892.1983.tb00368.x.

\bibitem[Rosenblatt(1959)]{Rosenblatt:59}
M.~Rosenblatt.
\newblock {Stationary Processes as Shifts of Functions of Independent Random
  Variables}.
\newblock \emph{{Journal of Mathematics and Mechanics}}, 8\penalty0
  (5):\penalty0 665--681, 1959.

\bibitem[Salinas et~al.(2019{\natexlab{a}})Salinas, Bohlke-Schneider, Callot,
  Medico, and Gasthaus]{salinas_high-dimensional_2019}
D.~Salinas, M.~Bohlke-Schneider, L.~Callot, R.~Medico, and J.~Gasthaus.
\newblock High-dimensional multivariate forecasting with low-rank {Gaussian}
  {Copula} {Processes}.
\newblock In \emph{Advances in {Neural} {Information} {Processing} {Systems}},
  volume~32. Curran Associates, Inc., 2019{\natexlab{a}}.
\newblock URL
  \url{https://proceedings.neurips.cc/paper/2019/hash/0b105cf1504c4e241fcc6d519ea962fb-Abstract.html}.

\bibitem[Salinas et~al.(2019{\natexlab{b}})Salinas, Flunkert, and
  Gasthaus]{salinas_deepar_2019}
D.~Salinas, V.~Flunkert, and J.~Gasthaus.
\newblock {DeepAR}: {Probabilistic} {Forecasting} with {Autoregressive}
  {Recurrent} {Networks}, Feb. 2019{\natexlab{b}}.
\newblock URL \url{http://arxiv.org/abs/1704.04110}.
\newblock arXiv:1704.04110 [cs, stat].

\bibitem[Schlegl et~al.(2019)Schlegl, Seeböck, Waldstein, Langs, and
  Schmidt-Erfurth]{Schlegl&Seebock:19}
T.~Schlegl, P.~Seeböck, S.~M. Waldstein, G.~Langs, and U.~Schmidt-Erfurth.
\newblock f-anogan: Fast unsupervised anomaly detection with generative
  adversarial networks.
\newblock \emph{Medical Image Analysis}, 54:\penalty0 30 -- 44, 2019.
\newblock ISSN 1361-8415.
\newblock \doi{https://doi.org/10.1016/j.media.2019.01.010}.

\bibitem[Tjostheim and Auestad(1994)]{tjostheim_nonparametric_1994}
D.~Tjostheim and B.~H. Auestad.
\newblock Nonparametric {Identification} of {Nonlinear} {Time} {Series}:
  {Projections}.
\newblock \emph{Journal of the American Statistical Association}, 89\penalty0
  (428):\penalty0 1398--1409, 1994.
\newblock ISSN 0162-1459.
\newblock \doi{10.2307/2291002}.
\newblock URL \url{https://www.jstor.org/stable/2291002}.
\newblock Publisher: [American Statistical Association, Taylor \& Francis,
  Ltd.].

\bibitem[Tong(1983)]{tong_threshold_1983}
H.~Tong.
\newblock \emph{Threshold {Models} in {Non}-linear {Time} {Series} {Analysis}},
  volume~21 of \emph{Lecture {Notes} in {Statistics}}.
\newblock Springer, New York, NY, 1983.
\newblock ISBN 978-0-387-90918-9 978-1-4684-7888-4.
\newblock \doi{10.1007/978-1-4684-7888-4}.
\newblock URL \url{http://link.springer.com/10.1007/978-1-4684-7888-4}.

\bibitem[Villani(2009)]{Villani09:Book}
C.~Villani.
\newblock \emph{The Wasserstein Distances}.
\newblock Springer Berlin Heidelberg, Berlin, Heidelberg, 2009.
\newblock ISBN 978-3-540-71050-9.
\newblock \doi{10.1007/978-3-540-71050-9_6}.
\newblock URL \url{https://doi.org/10.1007/978-3-540-71050-9_6}.

\bibitem[{Waibel} et~al.(1989){Waibel}, {Hanazawa}, {Hinton}, {Shikano}, and
  {Lang}]{Waibel&etal:89}
A.~{Waibel}, T.~{Hanazawa}, G.~{Hinton}, K.~{Shikano}, and K.~J. {Lang}.
\newblock {Phoneme Recognition Using Time-delay Neural Networks}.
\newblock \emph{IEEE Transactions on Acoustics, Speech, and Signal Processing},
  37\penalty0 (3):\penalty0 328--339, 1989.
\newblock \doi{10.1109/29.21701}.

\bibitem[Wang and Tong(2022)]{WangTong:21JMLR}
X.~Wang and L.~Tong.
\newblock {Innovations Autoencoder and its Application in One-class Anomalous
  Sequence Detection}.
\newblock \emph{Journal of Machine Learning Research}, 23\penalty0
  (49):\penalty0 1--27, 2022.
\newblock URL \url{http://jmlr.org/papers/v23/21-0735.html}.

\bibitem[Wang et~al.(2019)Wang, Smola, Maddix, Gasthaus, Foster, and
  Januschowski]{wang_deep_2019}
Y.~Wang, A.~Smola, D.~C. Maddix, J.~Gasthaus, D.~Foster, and T.~Januschowski.
\newblock Deep {Factors} for {Forecasting}, May 2019.
\newblock URL \url{http://arxiv.org/abs/1905.12417}.
\newblock arXiv:1905.12417 [cs, stat].

\bibitem[Weron(2014)]{weron_electricity_2014}
R.~Weron.
\newblock Electricity price forecasting: {A} review of the state-of-the-art
  with a look into the future.
\newblock \emph{International Journal of Forecasting}, 30\penalty0
  (4):\penalty0 1030--1081, Oct. 2014.
\newblock ISSN 0169-2070.
\newblock \doi{10.1016/j.ijforecast.2014.08.008}.
\newblock URL
  \url{https://www.sciencedirect.com/science/article/pii/S0169207014001083}.

\bibitem[Weron and Misiorek(2008)]{weron_forecasting_2008}
R.~Weron and A.~Misiorek.
\newblock Forecasting spot electricity prices: {A} comparison of parametric and
  semiparametric time series models.
\newblock \emph{International Journal of Forecasting}, 24\penalty0
  (4):\penalty0 744--763, 2008.
\newblock ISSN 0169-2070.
\newblock \doi{https://doi.org/10.1016/j.ijforecast.2008.08.004}.
\newblock URL
  \url{https://www.sciencedirect.com/science/article/pii/S0169207008000952}.

\bibitem[Wiener(1958)]{Wiener:58Book}
N.~Wiener.
\newblock \emph{{Nonlinear Problems in Random Theory}}.
\newblock Technology Press of Massachusetts Institute of Technology, Cambridge,
  MA, 1958.

\bibitem[Wu(2005)]{Wu:05PNAS}
W.~Wu.
\newblock {Nonlinear System Theory: Another Look at Dependence}.
\newblock \emph{Proceedings of National Academy of Sciences}, 102\penalty0
  (40):\penalty0 14150--14154, 2005.
\newblock URL \url{https://doi.org/10.1073/pnas.0506715102}.

\bibitem[Yoon et~al.(2022)Yoon, Park, Ryu, and Wang]{yoon_robust_2022}
T.~Yoon, Y.~Park, E.~K. Ryu, and Y.~Wang.
\newblock Robust {Probabilistic} {Time} {Series} {Forecasting}, Feb. 2022.
\newblock URL \url{http://arxiv.org/abs/2202.11910}.
\newblock arXiv:2202.11910 [cs].

\bibitem[Zhang and Yan(2023)]{zhang_crossformer_2023}
Y.~Zhang and J.~Yan.
\newblock Crossformer: {Transformer} {Utilizing} {Cross}-{Dimension}
  {Dependency} for {Multivariate} {Time} {Series} {Forecasting}.
\newblock Feb. 2023.
\newblock URL \url{https://openreview.net/forum?id=vSVLM2j9eie}.

\bibitem[Zhou et~al.(2021)Zhou, Zhang, Peng, Zhang, Li, Xiong, and
  Zhang]{zhou_informer_2021}
H.~Zhou, S.~Zhang, J.~Peng, S.~Zhang, J.~Li, H.~Xiong, and W.~Zhang.
\newblock Informer: {Beyond} {Efficient} {Transformer} for {Long} {Sequence}
  {Time}-{Series} {Forecasting}.
\newblock \emph{Proceedings of the AAAI Conference on Artificial Intelligence},
  35\penalty0 (12):\penalty0 11106--11115, May 2021.
\newblock ISSN 2374-3468.
\newblock \doi{10.1609/aaai.v35i12.17325}.
\newblock URL \url{https://ojs.aaai.org/index.php/AAAI/article/view/17325}.
\newblock Number: 12.

\bibitem[Zhou et~al.(2022)Zhou, Ma, Wen, Wang, Sun, and
  Jin]{zhou_fedformer_2022}
T.~Zhou, Z.~Ma, Q.~Wen, X.~Wang, L.~Sun, and R.~Jin.
\newblock {FEDformer}: {Frequency} {Enhanced} {Decomposed} {Transformer} for
  {Long}-term {Series} {Forecasting}, June 2022.
\newblock URL \url{http://arxiv.org/abs/2201.12740}.
\newblock arXiv:2201.12740 [cs, stat].

\end{thebibliography}

\newpage
\appendix

\section{Proof of Theorem \ref{thm:converge}}
For any fixed $n$, we define the following notation:
\begin{align*}
\nu_{t,m}^{*(n)} &= G_{\theta_m^*}(x_t,\cdots,x_{t-m+1}),\\
\hat{x}_{t,m}^{*(n)} &= H_{\eta_m^*}(\nu_{t,m}^{(n)},\cdots,\nu^{(n)}_{t-m+1,m}),\\
\tilde{\nu}_{t,m}^{(n)} &= G_{\tilde{\theta}_m}(x_t,\cdots,x_{t-m+1}).
\end{align*}
We denote the set of weights that obtain optimality under Eq.~\eqref{eq:loss} by $(\theta_m^*,\eta_m^*,\gamma_m^*,\omega_m^*)$. 
We make the clear distinction between the true weak innovations $\{\nu_t\}$ and estimated weak innovations with $m,n$-dimensional Weak Innovations Auto-encoder (WIAE), which is denoted by $\{\nu_{t,m}^{(n)}\}$.

Denote the $n$-dimensional vector $[\nu_{t,m}^{*(n)},\cdots,\nu_{t-n+1,m}^{*(n)}]$ by $\boldsymbol{\nu}_{t,m}^{*(n)}$, with the letter changed to bold face.
$\boldsymbol u_{t,m},\hat{\boldsymbol x}_{t,m}^{(n)}$ and $\boldsymbol x^{(n)}_{t,m}$ are defined similarly. $\{\tilde{\nu}_t\}$ is the sequence generated by $G_{\tilde{\theta}_m}$. All the random variables generated by $(\tilde{\theta}_m,\tilde{\eta}_m)$ are defined in the similar pattern.

{\em Proof:} 
\ben
\item Want to show that
$L_m^{(n)}(\tilde{\theta}_m,\tilde{\eta}_m)\rightarrow 0$ as $m\rightarrow \infty$.

By assumption A2, $G_{\tilde{\theta}_m}\rightarrow G$ uniformly, thus $\lVert \tilde{\boldsymbol\nu}_{t,m}^{(n)}-\boldsymbol\nu_t^{(n)}\rVert<n\epsilon$ for $\forall \epsilon>0$. 
Thus $\tilde{\boldsymbol\nu}_{t,m}^{(n)}\stackrel{d}{\Rightarrow}\boldsymbol\nu_t$. 
Similarly, we have $[H_{\tilde{\eta}_m}(\nu_{t},\cdots,\nu_{t-m+1}),\cdots,H_{\tilde{\eta}_m}(\nu_{t-n+1},\cdots,\nu_{t-n-m+2})]\stackrel{d}{\Rightarrow} [x_t,\cdots,x_{t-n+1}]$.

Since $H$ is continuous and $H_{\tilde{\eta}_m}$ converge uniformly to $H$, $H_{\tilde{\eta}_m}$  is continuous. Thus by continuous mapping theorem, $H_{\tilde{\eta}_m}(\tilde{\nu}_{t,m}^{(n)},\cdots,\tilde{\nu}_{t-m+1,m}^{(n)})\stackrel{d}{\Rightarrow}H_{\tilde{\eta}_m}(\nu_{t},\cdots,\nu_{t-m+1})$. Thus, 
\[[H_{\tilde{\eta}_m}(\tilde{\nu}_{t,m}^{(n)},\cdots,\tilde{\nu}_{t-m+1,m}^{(n)}),\cdots,H_{\tilde{\eta}_m}(\tilde{\nu}_{t-n+1,m}^{(n)},\cdots,\tilde{\nu}^{(n)}_{t-n-m+2,m})]\stackrel{d}{\Rightarrow} [x_t,\cdots,x_{t-n+1}]\] 
Hence $L_m^{(n)}(\tilde{\theta}_m,\tilde{\eta}_m)\rightarrow 0$.

Therefore, $L_m^{(n)}(\theta_m^*,\eta_m^*):=\min_{\theta,\eta}L^{(n)}_m(\theta,\eta) \leq L_m^{(n)}(\tilde{\theta}_m,\tilde{\eta}_m)\rightarrow 0$.

\item Since $L^{(n)}(\theta_m^*,\eta_m^*)\rightarrow 0$ as $m\rightarrow \infty$, by Eq.~\eqref{eq:loss}, $\boldsymbol\nu_{t,m}^{(n)} \stackrel{d}{\Rightarrow} \boldsymbol u_{t,m}^{(n)}$, $\hat{\boldsymbol x}_{t,m}^{(n)}\stackrel{d}{\Rightarrow}\boldsymbol x^{(n)}_{t,m}$ follow directly by the equivalence of convergence in Wasserstein distance and convergence in distribution \citep{Villani09:Book}.
\een

\section{Proof of Theorem \ref{thm:sufficiency}}
{\em Proof:} By the definition of weak innovations, $\{x_t\}\stackrel{d}{=}\{\hat{x}_t\}$. Hence we know that the conditional cumulative density function (CDF) of $\{\hat{x}_t\}$ satisfies
\begin{multline}
    \mathbb{P}[\hat{x}_t\leq x|\hat{x}_{t-1}=a_{t-1},\hat{x}_{t-2}=a_{t-2},\cdots]=\mathbb{P}[x_t\leq x|x_{t-1}=a_{t-1},x_{t-2}=a_{t-2},\cdots]\\=F(x|a_{t-1},a_{t-2},\cdots).
\end{multline}
Since the infinite dimensional decoder function $H$ is injective,
\[\mathbb{P}[\hat{x}_t\leq x|\hat{x}_{t-1}=a_{t-1},\hat{x}_{t-2}=a_{t-2},\cdots] = \mathbb{P}[\hat{x}_t\leq x|\nu_{t-1}=b_{t-1},\nu_{t-2}=b_{t-2},\cdots],\]
where $a_{s} = H(\nu_s,\nu_{s-1},\cdots)$.
Hence theorem \ref{thm:sufficiency} holds.

\section{Dataset Details for Performance Evaluation}
In Sec.~\ref{subsec:rep-learning}, we tested out method on three different synthetic datasets named as LAR, MAR and MC. 
The specific formulation of the synthetic datasets is shown in Table.~\ref{tab:Synthetic dataset}.
The three synthetic cases were designed to cover three different existence scenarios of the innovations representation.
The Linear Auto Regression (LAR) case has known innovations sequence \citep{WangTong:21JMLR} that can be extracted through linear function pair.
Consequently, the weak innovations sequence must exist for the LAR case.
 On the other hand, the MA case satisfies the existence condition for weak innovations sequence.
 Since the MA case is of minimum phase, whether it has innovations representation remains agnostic.   
 The two-state Markov Chain case (MC) with transition probability specified in Table.~\ref{tab:Synthetic dataset}, only has weak innovations as shown by \cite{Rosenblatt:59}.

\begin{table}[t]
\caption{Test Synthetic Datasets. $u_t\stackrel{\tiny\rm i.i.d}{\sim}\mathcal{U}[-1,1]$.}
\label{tab:Synthetic dataset}
\begin{center}
\begin{small}
\begin{sc}
\begin{tabular}{ll}
\toprule
Dataset & Model \\
\midrule
Moving Average (MA)    &$x_t=u_t+2.5u_{t-1}$  \\
Linear Autoregressive (LAR) &$x_t=0.5 x_{t-1}+u_t$ \\
Two-State Markov Chain (MC)    &$P=  \begin{bmatrix} 0.6 & 0.4\\0.4 &0.6\end{bmatrix}$\\
\bottomrule
\end{tabular}
\end{sc}
\end{small}
\end{center}
\vspace{-2em}
\end{table}

\section{Definition of Metrics}
\label{sec:metrics}
In this section, we present the formulas of metrics used in the simulation section.
Let $\{x_t\}$ denote the original times series, $\{\hat{x}_t\}$ the prediction estimates, $N$ the size of datasets and $s$ the prediction step.
\begin{align}
    NMSE = \frac{\frac{1}{N}\sum_{t=1}^N(x_t-\hat{x}_t)^2}{\frac{1}{N}\sum_{t=1}^N x_t^2},\\
    NMeSE = med\left(\left(\frac{(x_t-\hat{x}_t)^2}{\frac{1}{N}\sum_{t=1}^N x_t^2}\right)_t\right),\\
    NMAE = \frac{\frac{1}{N}\sum_{t=1}^N|x_t-\hat{x}_t|}{\frac{1}{N}\sum_{t=1}^N|x_t|},\\
    NMeAE = med\left(\left(\frac{|x_t-\hat{x}_t|}{\frac{1}{N}\sum_{t=1}^N|x_t|}\right)_t\right),\\
    MASE = \frac{\frac{1}{N}\sum_{t=1}^N|x_t-\hat{x}_t|}{\frac{1}{N-s}\sum_{t=s+1}^N|x_t-x_{t-s}|},\\
    sMAPE = \frac{1}{N}\sum_{t=1}^N\frac{|x_t-\hat{x}_t|}{(|x_t|+|\hat{x}_t|)/2},
\end{align}
where $med(\cdot)$ denotes the operation of taking median of a given sequence.
The purpose of adopting multiple metrics is to comprehensively evaluate the level of performance for each method.
Especially for electricity price datasets with high variability, the methods with best median errors and best mean errors could be very different.
Mean errors, i.e., NMSE and NMAE, represent the overall performance, but are not robust to outliers.
On the other hand, median errors (NMeSE and NMeAE) are robust to outliers. 
MASE reflects the relative performance to the naive forecaster, which utilizes past samples as forecast.
Methods with MASE smaller than $1$ outperform the naive forecaster.
sMAPE is the symmetric counterpart of mean absolute percentage error (MAPE) that can be both upper bounded and lower bounded.
Since for electricity datasets, the actual values can be very close to $0$, thus nullifies the effectiveness of MAPE, we though that sMAPE is the better metric.

\section{Implementation Details of WIAE and Benchmarks}
DeepAR, NPTS and Wavenet are constructed from GluonTS \footnote{\url{https://ts.gluon.ai/stable/\#}} time series forecasting package.
We implemented WIAE, TLAE, IAE, Anica and fAnoGAN by ourselves. \footnote{WIAE can be found on the using the github link: \url{https://github.com/Lambelle/WIAE.git}.} for comparison. 
All the neural networks (encoder, decoder and discriminator) in the paper had three hidden layers, with the 100, 50, 25 neurons respectively. In the paper, $m=20, n=50$ was used for all cases. The encoder and decoder both used hyperbolic tangent activation. The first two layers of the discriminator adopted hyperbolic tangent activation, and the last one linear activation.

In training we use Adam optimizer with $\beta_1=0.9$, $\beta_2=0.999$. Batch size and epoch are set to be $60$ and $100$, respectively. The detailed hyperparameter of choice can be found in Table.~\ref{tb:Hyperparameters}. We note that under different version of packages and different cpus, the result might be different (even if we set random seed). The experimental results are obtained using dependencies specified in Github repository\footnote{\url{https://github.com/Lambelle/WIAE.git}}.
\begin{table}[t]
\caption{Hyper Parameters Setting for Each Case}
\label{tb:Hyperparameters}
\begin{center}
\begin{small}
\begin{sc}
\begin{tabular}{lcccl}
\toprule
\textbf{Test Case} & \textbf{Learning Rate} $\mathbf{\alpha}$ & \textbf{Gradient Penalty} $\mathbf{\lambda_1,\lambda_2}$ &\textbf{Weight for Reconstruction} ($\lambda$) \\
\midrule
MC   &$0.0001$ &$1.0,1.0$ &$1.0$  \\
AR1    &$0.0001$   &$1.0,1.6$ &$1.0$ \\
MA     &$0.0001$  &$1.0,1.6$ &$1.0$\\
SP500   &$1e-5$ &$1.0,1.3$  &$1.0$\\
NYISO   &$1e-5$    &$1.0,1.4$ &$1.0$\\
ISONE   &$1e-5$     &$1.0,1.0$ &$1.0$\\
PJM     &$1e-5$ &$1.0,1.0$ &$1.0$\\
electricity &$1e-5$ &$1.0,1.0$ &$1.0$\\
Traffic   &$1e-5$ &$1.0,1.2$ &$1.0$\\
\bottomrule
\end{tabular}
\end{sc}
\end{small}
\end{center}
\end{table}

\section{Additional Time series Forecasting Results}
\begin{table}[t]
  \caption{Numerical Results Continued. The numbers in the parenthese indicate the prediction step.}
  \label{tb:Result-2}
  \centering
  \begin{tabular}{lllllllllll}
    \toprule
    Metrics &Method     & Electricity($5$)    & Electricity($10$) & Traffic($8$) &Traffic($20$) &PJM($5$) &PJM($12$) \\
    \midrule
       \multirow{6}{*}{\textbf{NMSE}} &WIAE &$0.5165$  & $0.5165$ &$0.1091$ &$0.1299$  &$1.9013$ &$0.13509$   \\
    &DeepAR    &$0.5290$ &$0.6272$ &$0.0309$ &$0.0797$ &$\mathbf{0.4401}$ &$\mathbf{0.3893}$ \\
    &NPTS     &$1.3862$ &$1.3867$ &$0.3045$ &$0.3050$ & $8.5330$ &$8.5555$  \\
    &Pyraformer &$\mathbf{0.1422}$ &$\mathbf{0.1735}$ &$0.2722$ &$0.3744$ &$1.3511$ &$1.5476$\\
    &TLAE &$0.1881$ &$0.2304$ &$0.1097$ &$0.2790$ &$0.5210$ &$0.6470$\\
    &Wavenet &$1.3367$ &$1.3575$ &$\mathbf{0.0211}$ &$\mathbf{0.0241}$ &$0.4430$ &$0.4666$\\
    \midrule
    \multirow{6}{*}{\textbf{NMeSE}}
    &WIAE &$0.1416$  & $0.1626$ &$0.0237$ &$0.0578$ &$0.0156$  &$0.0281$   \\
    &DeepAR    &$0.1280$ &$0.1500$ &$\mathbf{0.0046}$ &$0.0200$ &$0.0164$ &$0.0196$ \\
    &NPTS     &$0.9135$ &$0.9136$ &$0.0893$ &$0.0894$ & $0.0241$ &$0.0245$   \\
    &Pyraformer &$0.0139$ &$0.0181$ &$0.1194$ &$0.1830$ &$0.0808$ &$0.1032$\\
    &TLAE &$\mathbf{0.0135}$ &$\mathbf{0.0151}$ &$0.0261$ &$0.1508$ &$\mathbf{0.0143}$ &$\mathbf{0.0190}$\\
    &Wavenet &$1.4559$ &$1.4895$ &$0.0211$ &$\mathbf{0.0036}$ &$0.0182$ &$0.0225$\\
    \midrule
    \multirow{6}{*}{\textbf{NMAE}}&WIAE &$0.3296$  & $\mathbf{0.3487}$ &$0.2350$ &$0.3012$  &$0.3191$ &$0.4586$   \\
    &DeepAR    &$0.3360$ &$0.3559$ &$0.1144$ &$0.2113$ &$\mathbf{0.2285}$ &$\mathbf{0.2475}$ \\
    &NPTS     &$0.4625$ &$0.4627$ &$0.4060$ &$0.4065$ & $0.5512$ &$0.5525$   \\
    &Pyraformer &$0.3396$ &$0.3794$ &$0.4544$ &$0.5293$ &$0.4819$ &$0.5534$\\
    &TLAE  &$\mathbf{0.3041}$ &$0.3597$ &$0.2710$ &$0.4641$ &$0.1804$ &$0.2059$\\
    &Wavenet  &$0.6215$ &$0.6216$ &$\mathbf{0.0864}$ &$\mathbf{0.0940}$ &$0.2605$ &$0.2844$\\
    \midrule
    \multirow{6}{*}{\textbf{NMeAE}}
    &WIAE &$0.2619$  & $0.2461$ &$0.1602$ &$0.2607$  &$\mathbf{0.1175}$ &$0.1521$   \\
    &DeepAR    &$0.2332$ &$0.2533$ &$0.0738$ &$0.1562$ &$0.1236$ &$0.1333$    \\
    &NPTS     &$0.6190$ &$0.6191$ &$0.3244$ &$0.3245$ &$0.1461$ &$0.1464$\\
    &Pyraformer &$0.1755$ &$0.2003$ &$0.3754$ &$0.4663$ & $0.2206$ &$0.2304$   \\
    &TLAE  &$\mathbf{0.1571}$ &$\mathbf{0.1729}$ &$0.1793$ &$0.4314$ &$0.2206$ &$0.2304$\\
    &Wavenet &$0.7084$ &$0.7085$ &$\mathbf{0.0596}$ &$\mathbf{0.0627}$ &$0.1205$ &$\mathbf{0.1408}$\\
    \midrule
    \multirow{6}{*}{\textbf{MASE}}
    &WIAE &$0.8309$  & $\mathbf{0.8565}$ &$0.7948$  &$0.9208$  &$1.1307$ &$0.9810$ \\
    &DeepAR    &$0.8721$ &$0.9717$ &$0.5884$ &$0.5712$ &$\mathbf{0.8426}$ &$\mathbf{0.8245}$   \\
    &NPTS     &$1.996$ &$1.1151$ &$1.8938$ &$1.0654$ & $1.6061$ &$1.2690$   \\
    &Pyraformer  &$1.0897$ &$1.0723$ &$1.9028$ &$1.3451$ &$1.4990$ &$1.1382$\\
    &TLAE   &$\mathbf{0.8002}$ &$0.9865$ &$0.9999$ &$0.9999$ &$0.9979$ &$0.9765$ \\
    &Wavenet &$2.6178$ &$2.4475$ &$\mathbf{0.4913}$ &$\mathbf{0.2945}$ &$1.9064$ &$0.8932$\\
    \midrule
    \multirow{6}{*}{\textbf{sMAPE}}
    &WIAE &$0.5274$  & $0.5034$ &$0.2861$ &$0.3715$  &$0.2754$  &$0.4161$ \\
    &DeepAR    &$0.4917$ &$0.5004$ &$0.1365$ &$0.2574$ &$0.2743$ &$0.3101$    \\
    &NPTS     &$0.6051$ &$0.6053$ &$0.4871$ &$0.4876$ & $0.8932$ &$0.8934$  \\
    &Pyraformer  &$0.6385$ &$0.6546$ &$0.5489$ &$0.6264$ &$0.3968$ &$0.4039$\\
    &TLAE   &$\mathbf{0.3974}$ &$\mathbf{0.4791}$ &$0.3184$ &$0.4995$ &$\mathbf{0.1934}$ &$\mathbf{0.2020}$\\
    &Wavenet &$1.9877$ &$1.9858$ &$\mathbf{0.1069}$ &$\mathbf{0.1164}$ &$0.3182$ &$0.3382$\\
    \bottomrule
  \end{tabular}
\end{table}

Due to page limitation, we show additional experimental results in this section.
Table.~\ref{tb:Result-2} shows the results for PJM, {\it traffic} and {\it electricity} datasets.
{\it Traffic} and {\it electricity} datasets are rather smooth measurements collected from physical systems, where Wavenet, the auto-regressive model performed relatively well.
PJM dataset contains hourly sampled electricity prices that are hard to predict. 
For different metrics, different methods performs well.

\section{Empirical CDF of Errors}

\begin{figure}[t]
    \centering
    \includegraphics[scale=0.25]{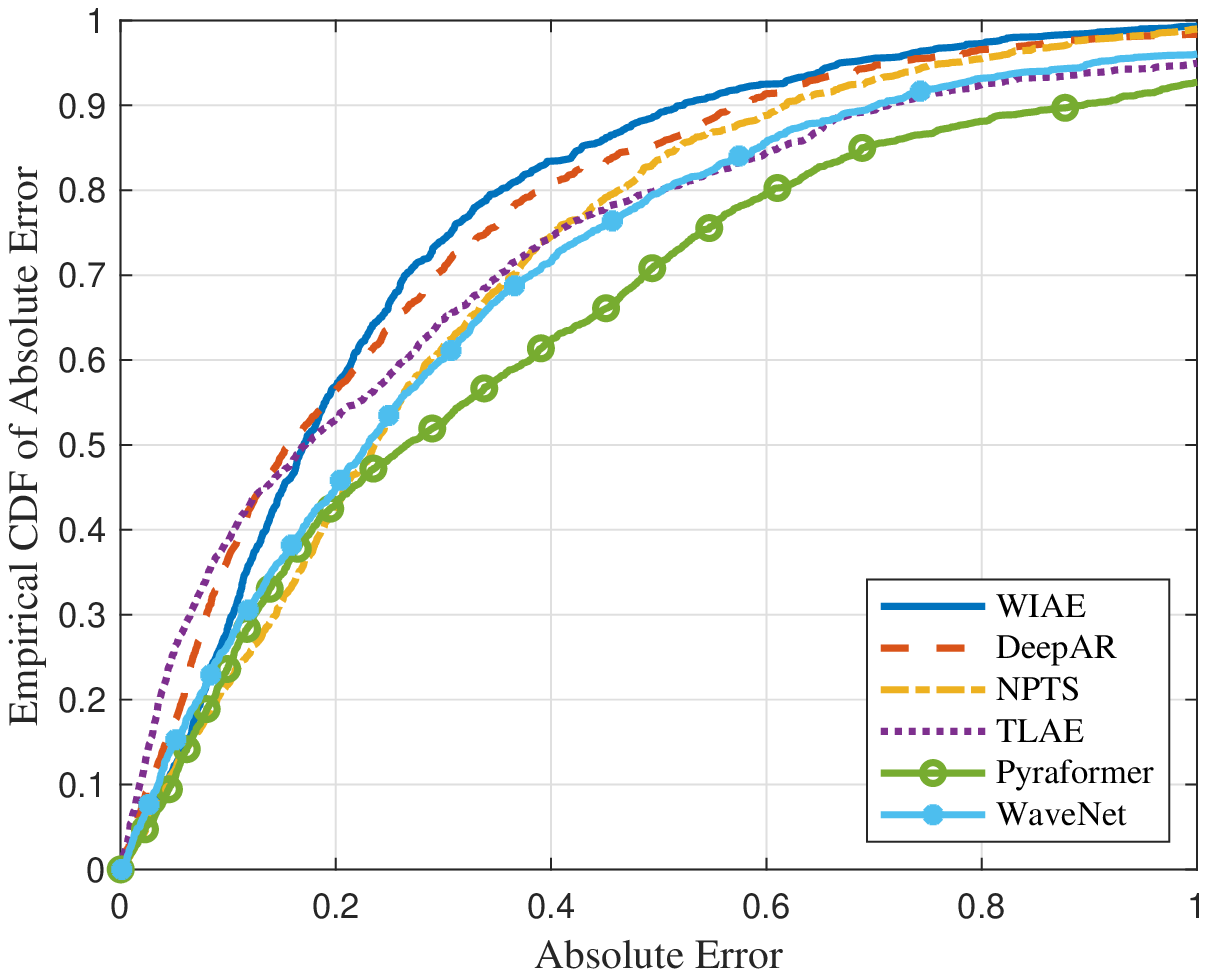}
    \includegraphics[scale=0.25]{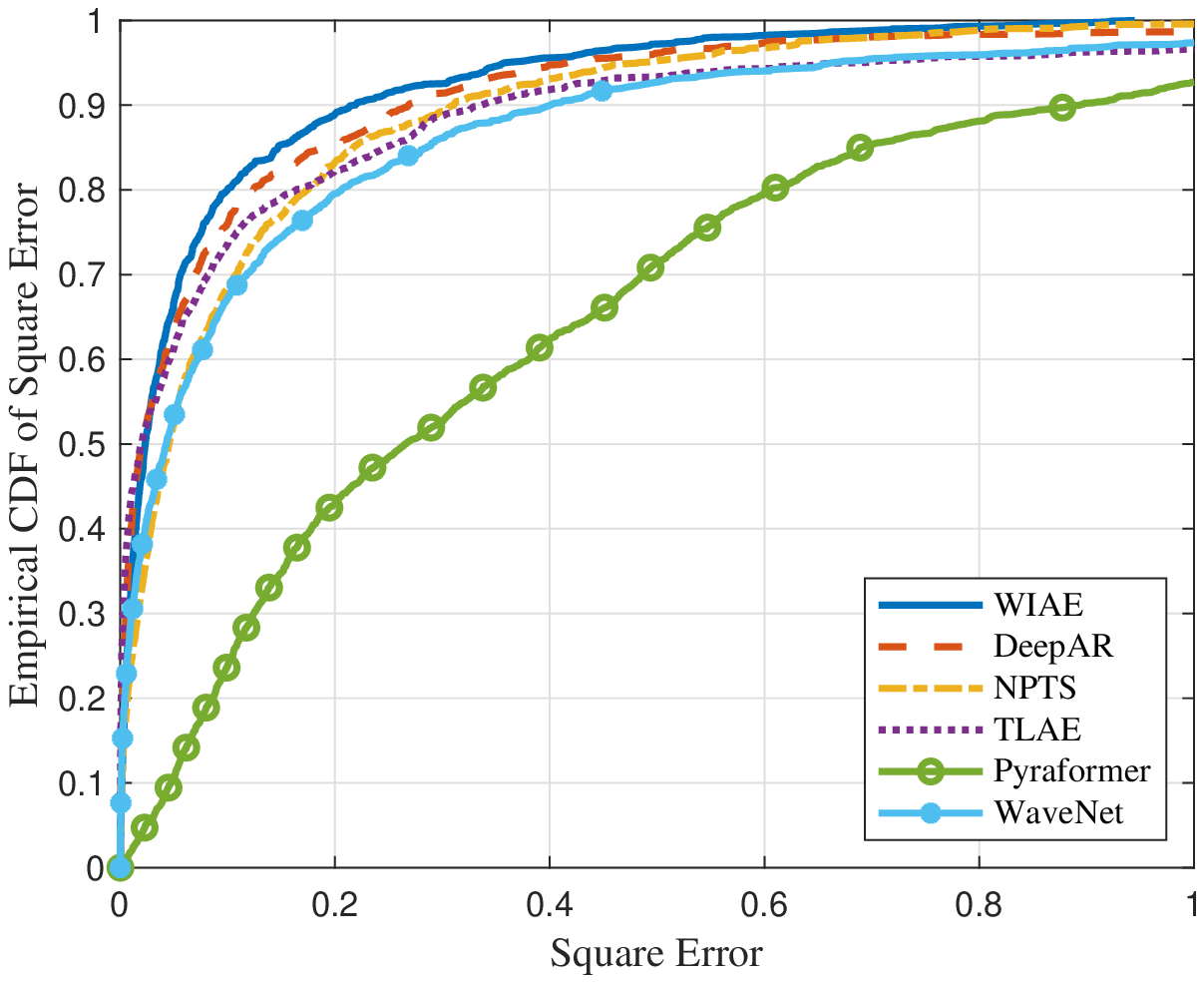}
    \includegraphics[scale=0.25]{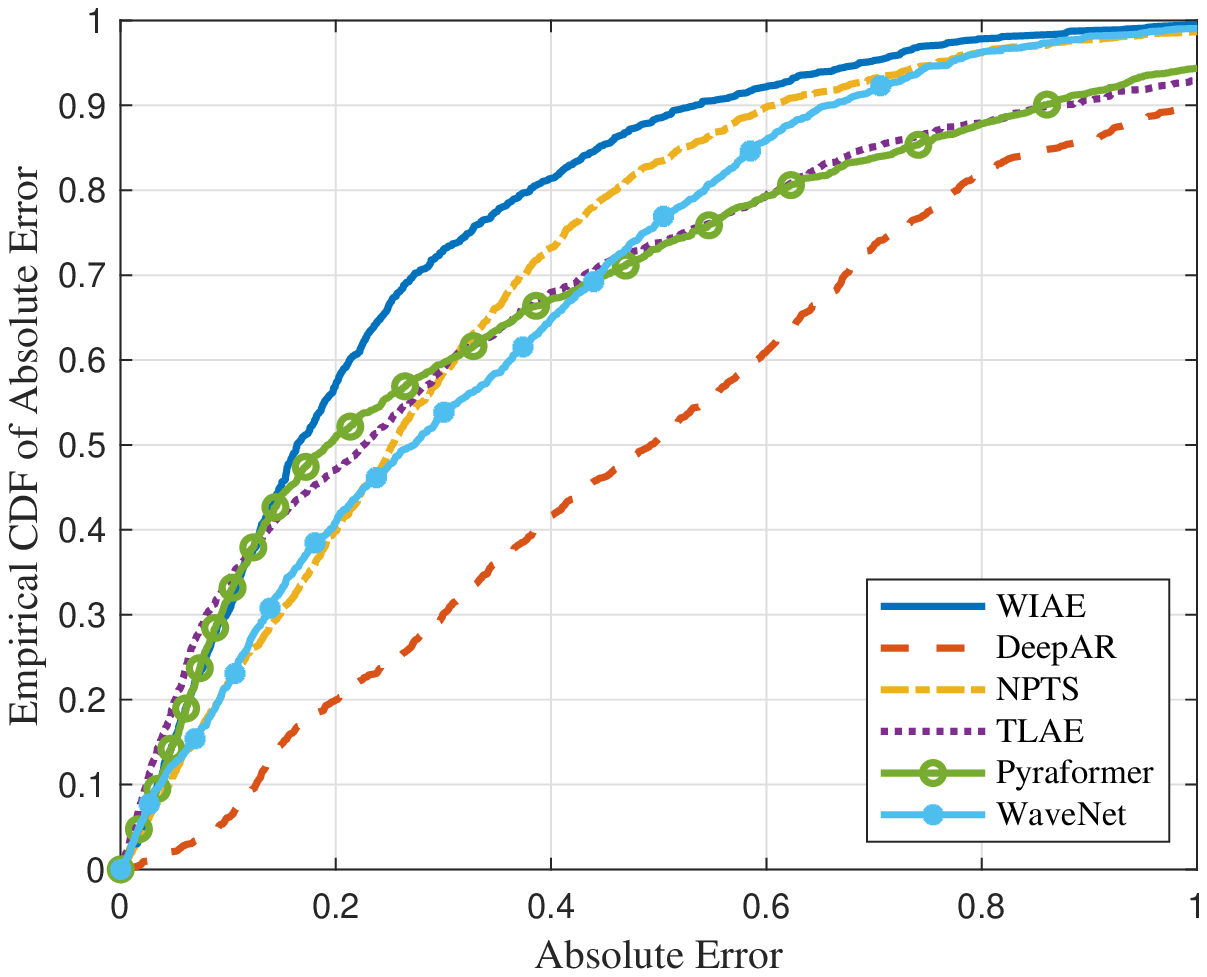}
    \includegraphics[scale=0.25]{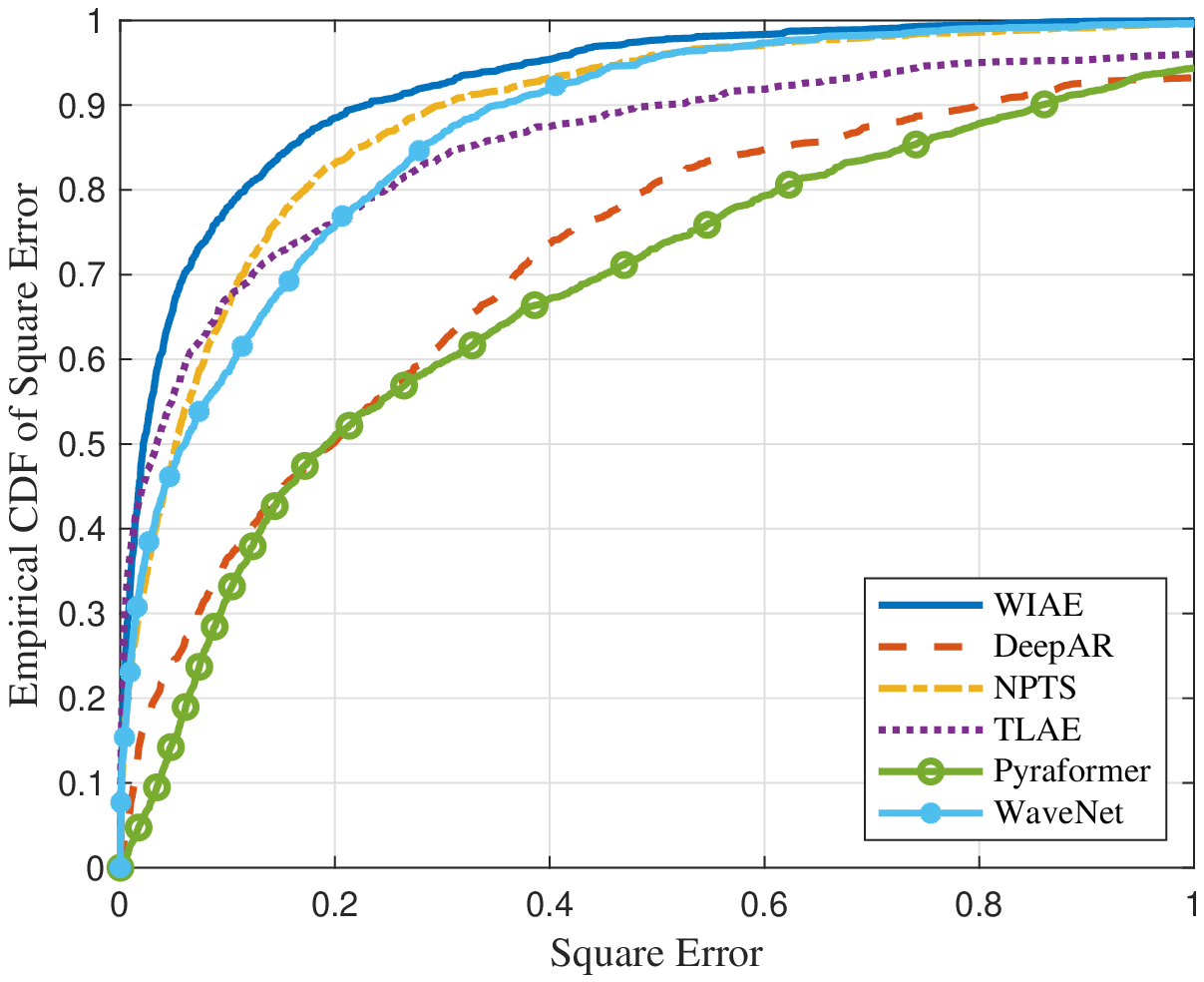}
    \caption{Errors for ISONE 15-step (two subplots to the left) and 24-step (two subplots to the right) prediction.}
    \label{fig:ecdf_isone}
\end{figure}

\begin{figure}[t]
    \centering
    \includegraphics[scale=0.25]{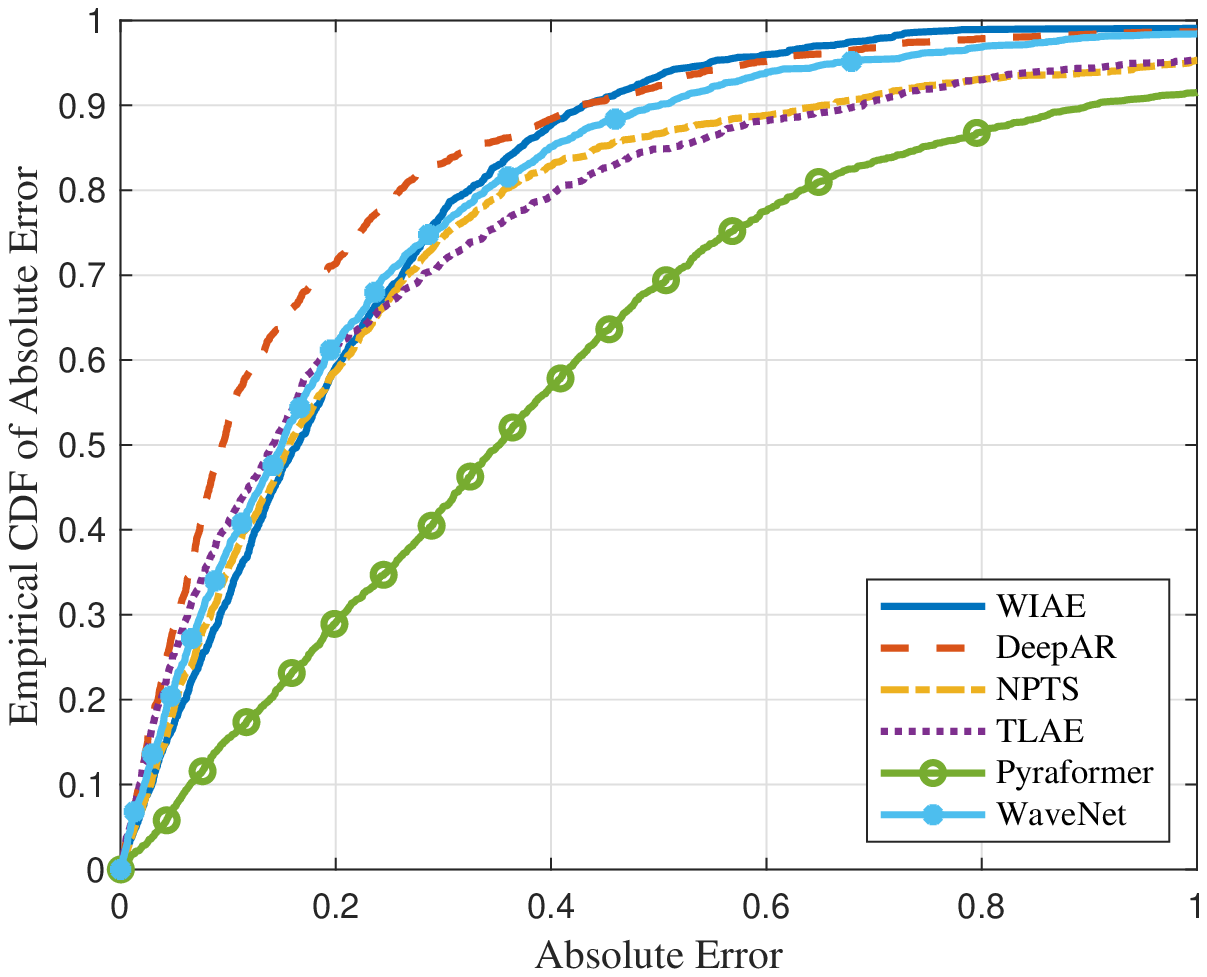}
    \includegraphics[scale=0.25]{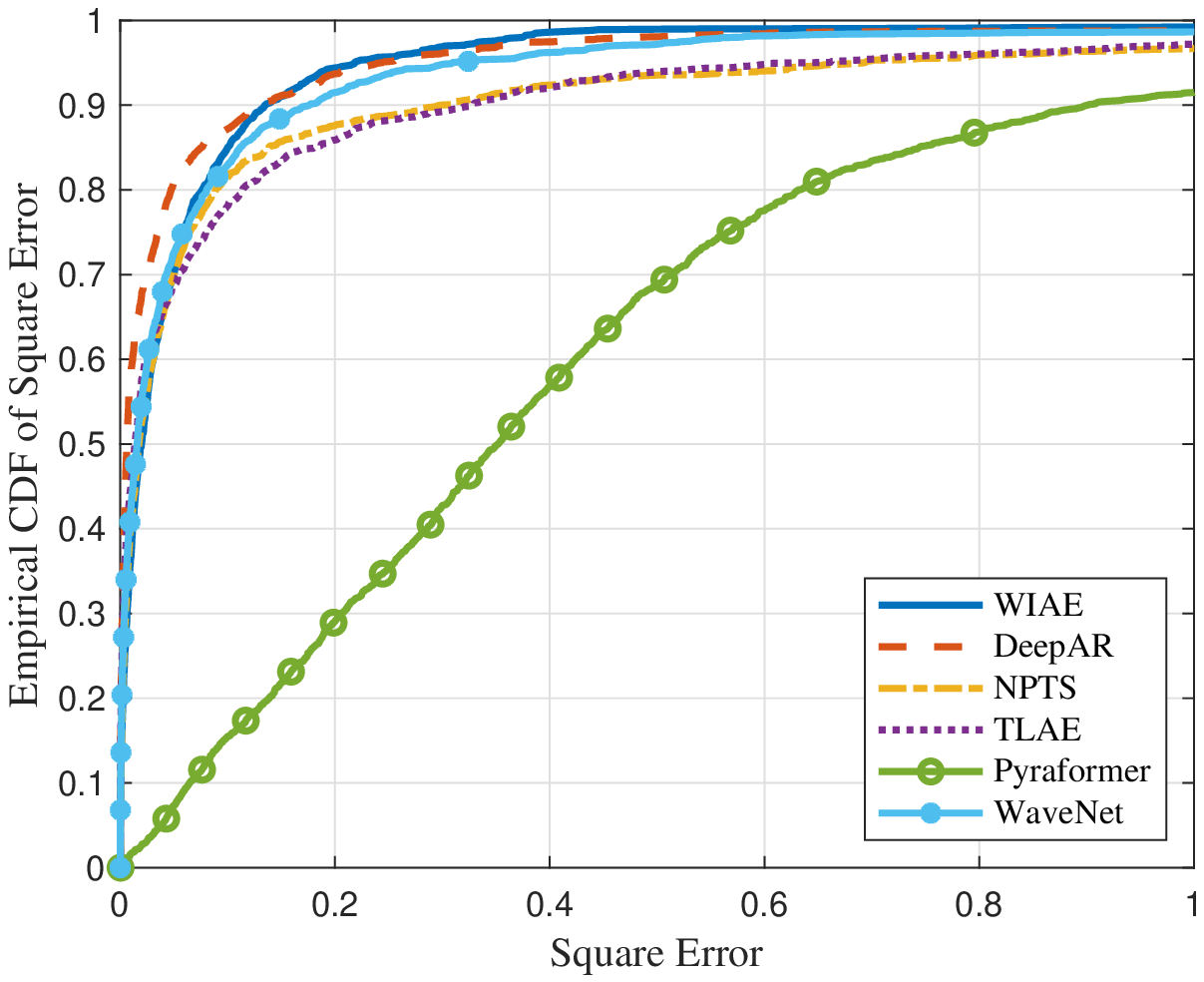}
    \includegraphics[scale=0.25]{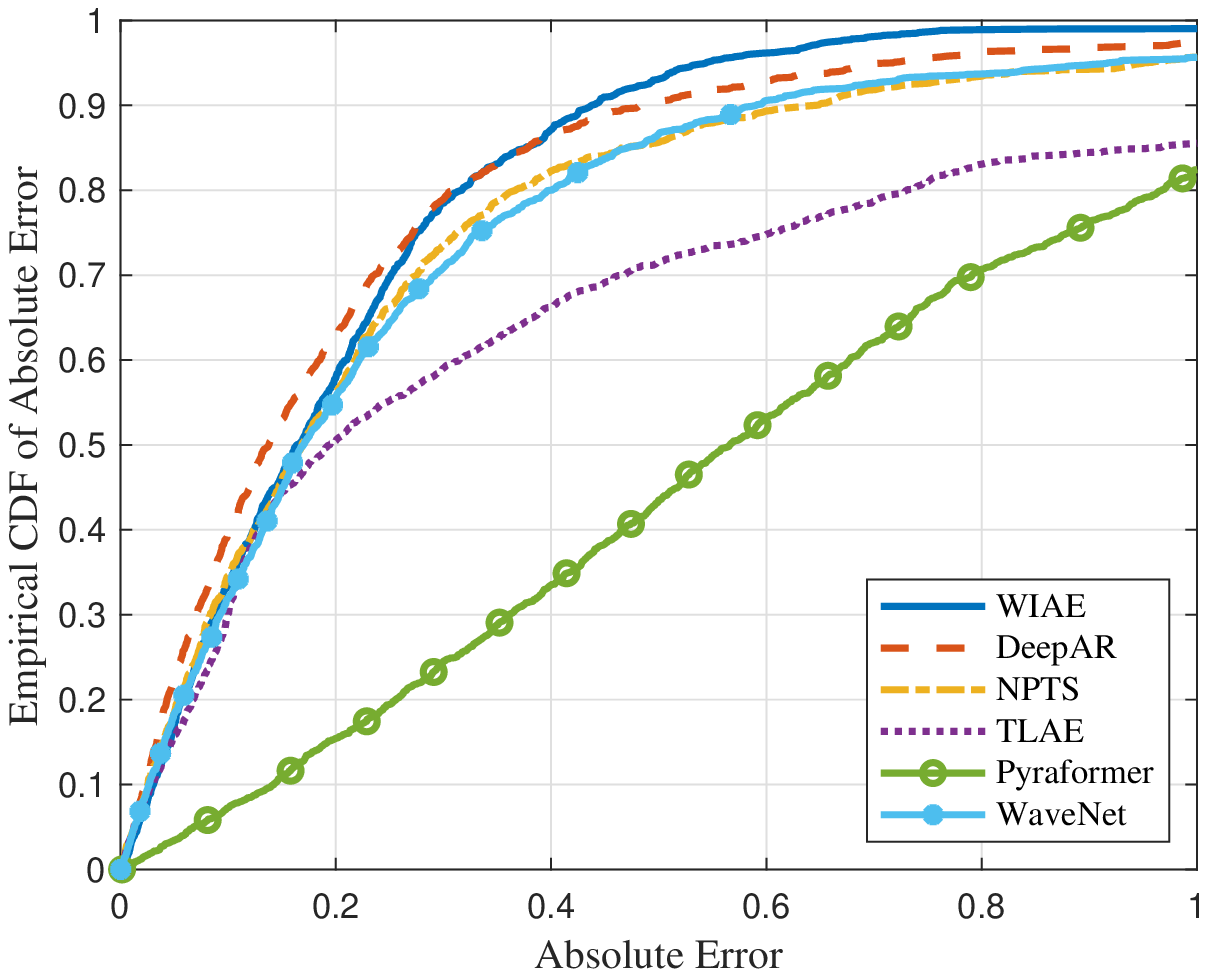}
    \includegraphics[scale=0.25]{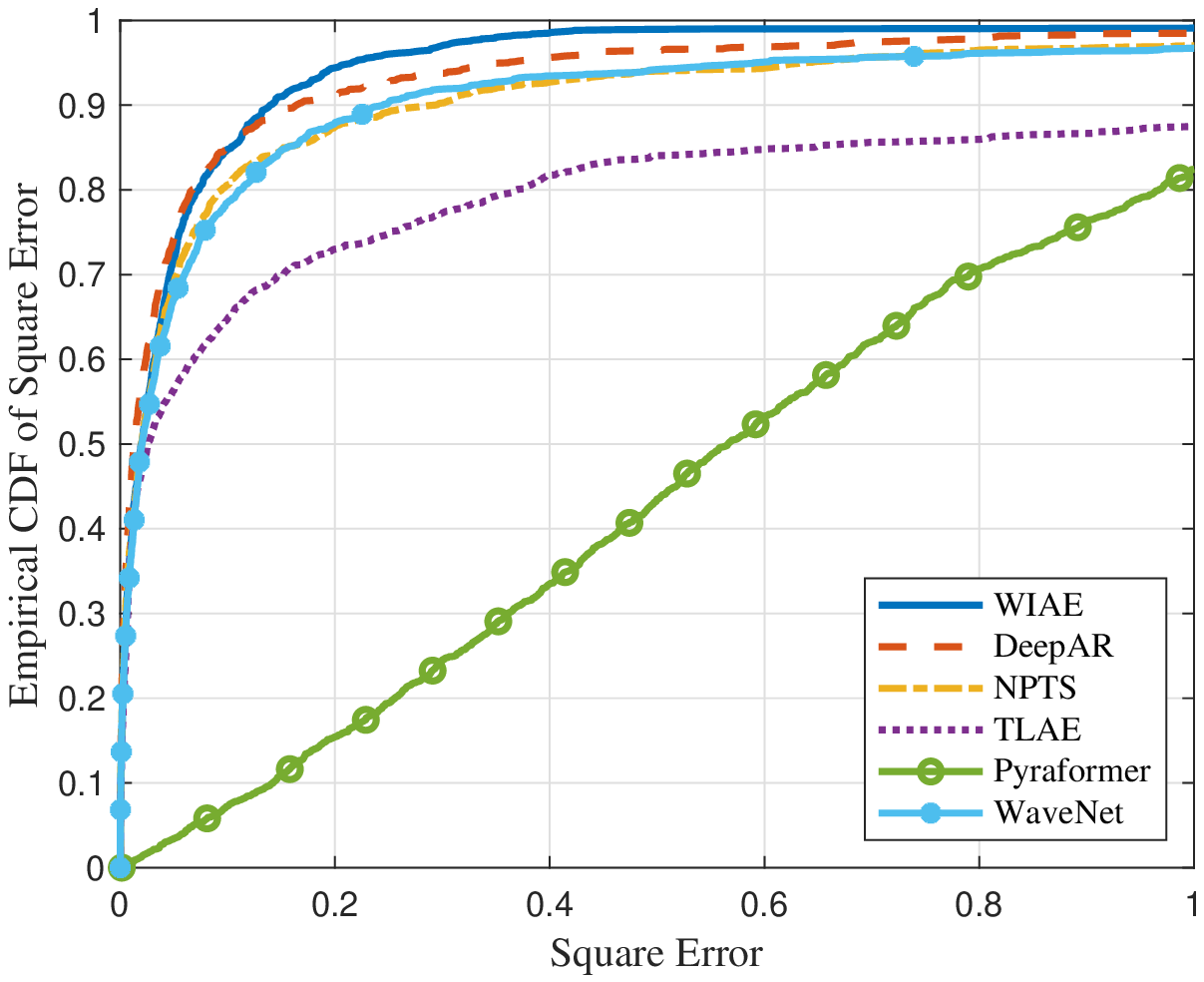}
    \caption{Errors for NYISO 15-step (two subplots to the left) and 24-step (two subplots to the right) prediction.}
    \label{fig:ecdf_nyiso}
\end{figure}

\begin{figure}[t]
    \centering
    \includegraphics[scale=0.25]{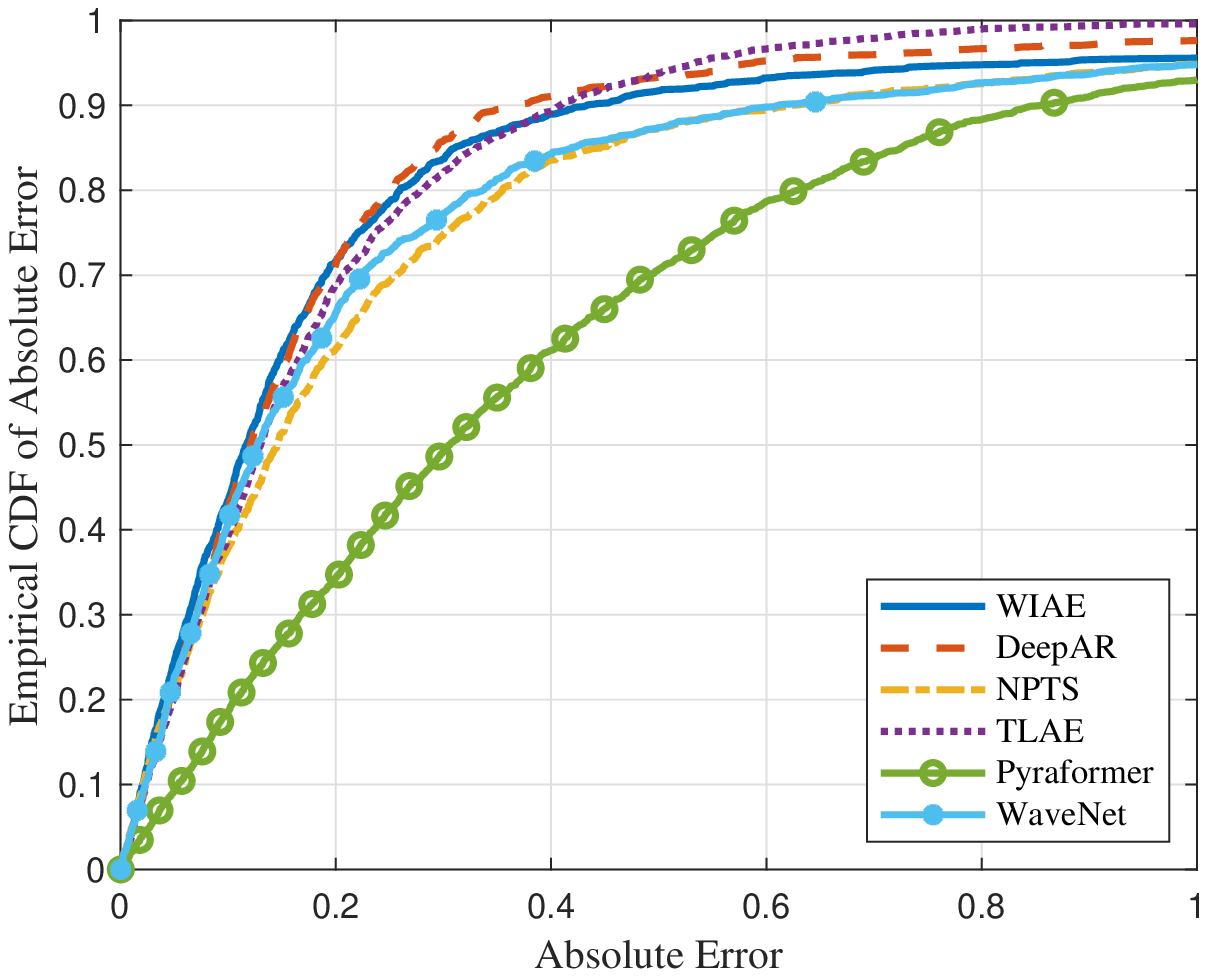}
    \includegraphics[scale=0.25]{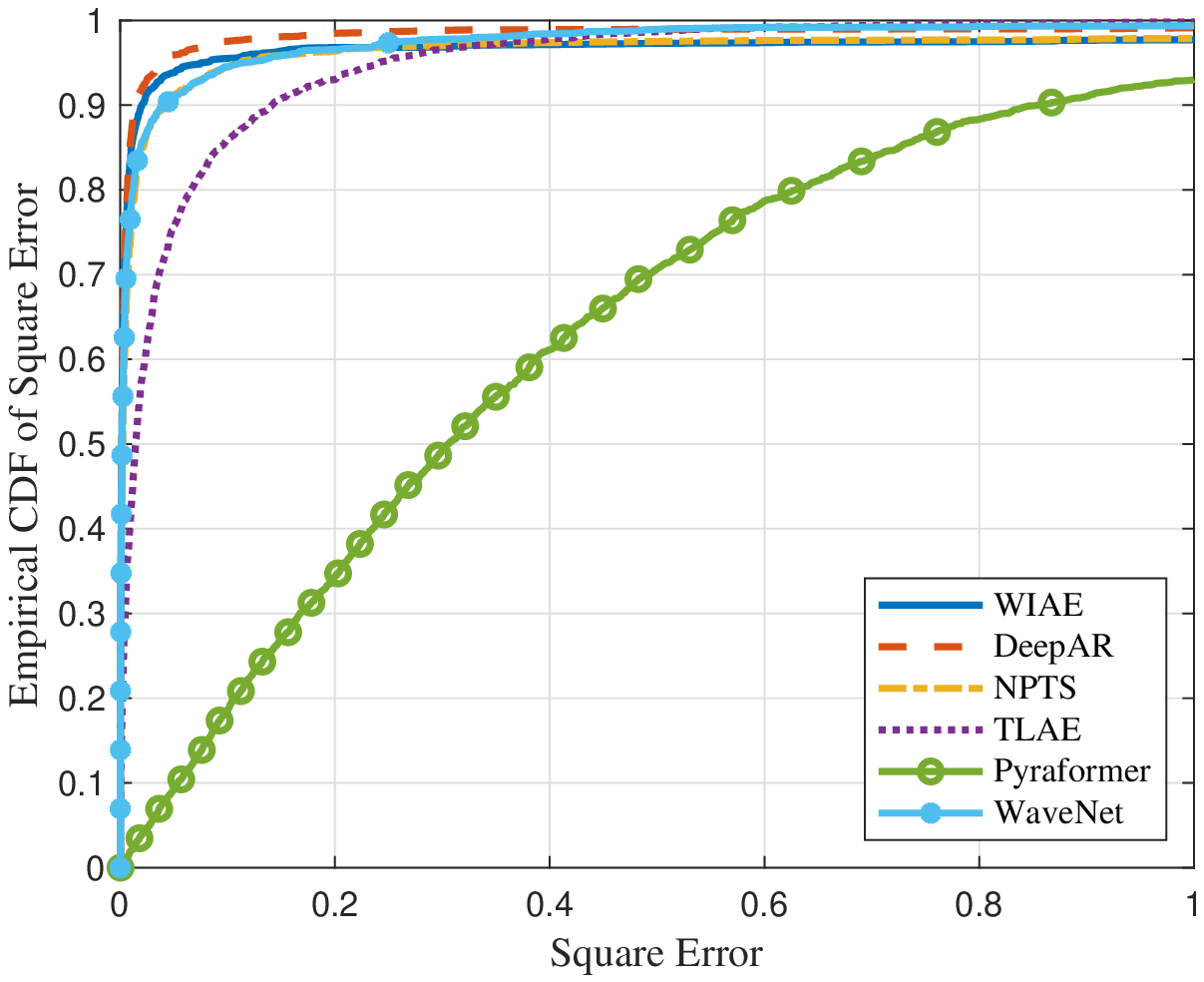}
    \includegraphics[scale=0.25]{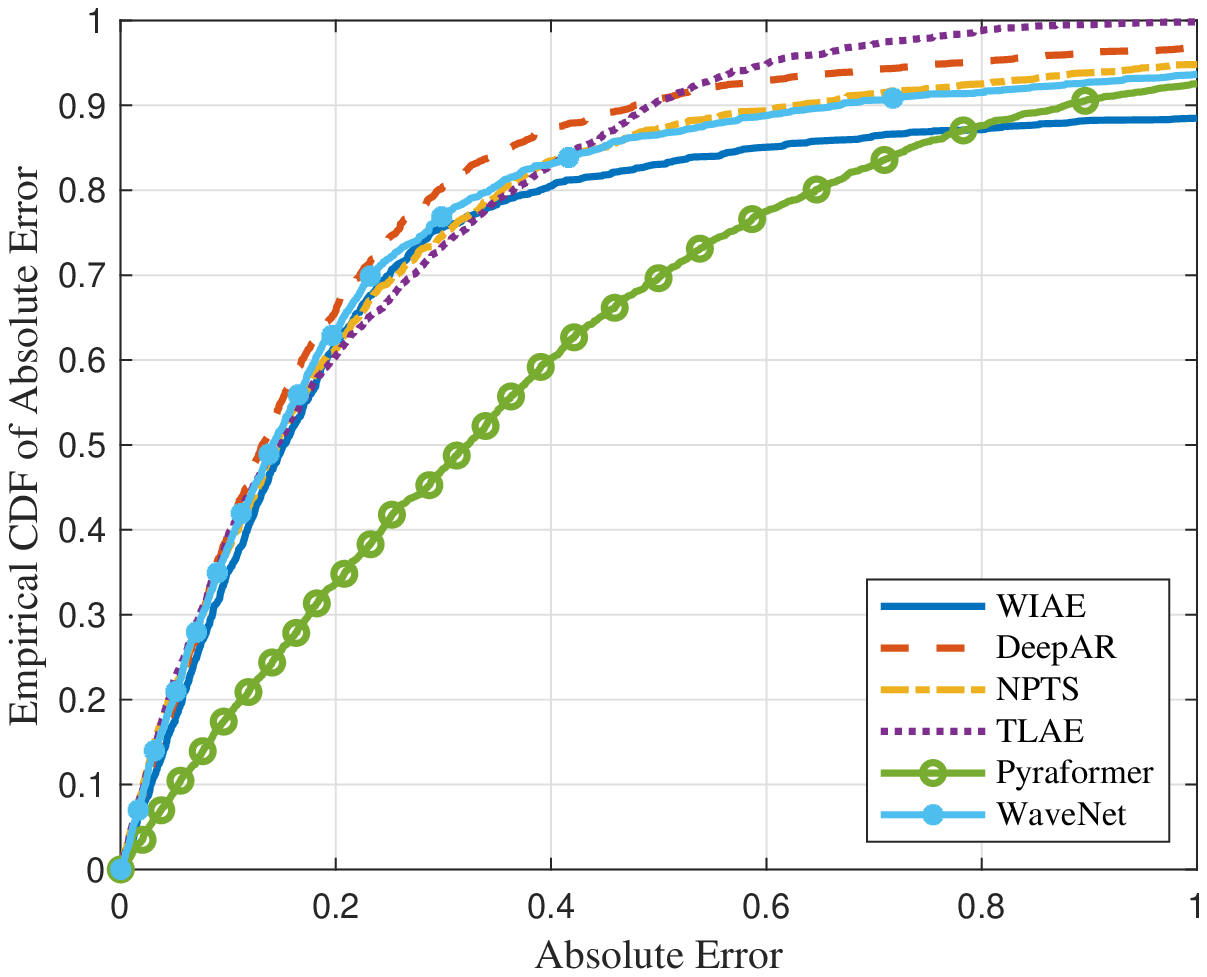}
    \includegraphics[scale=0.25]{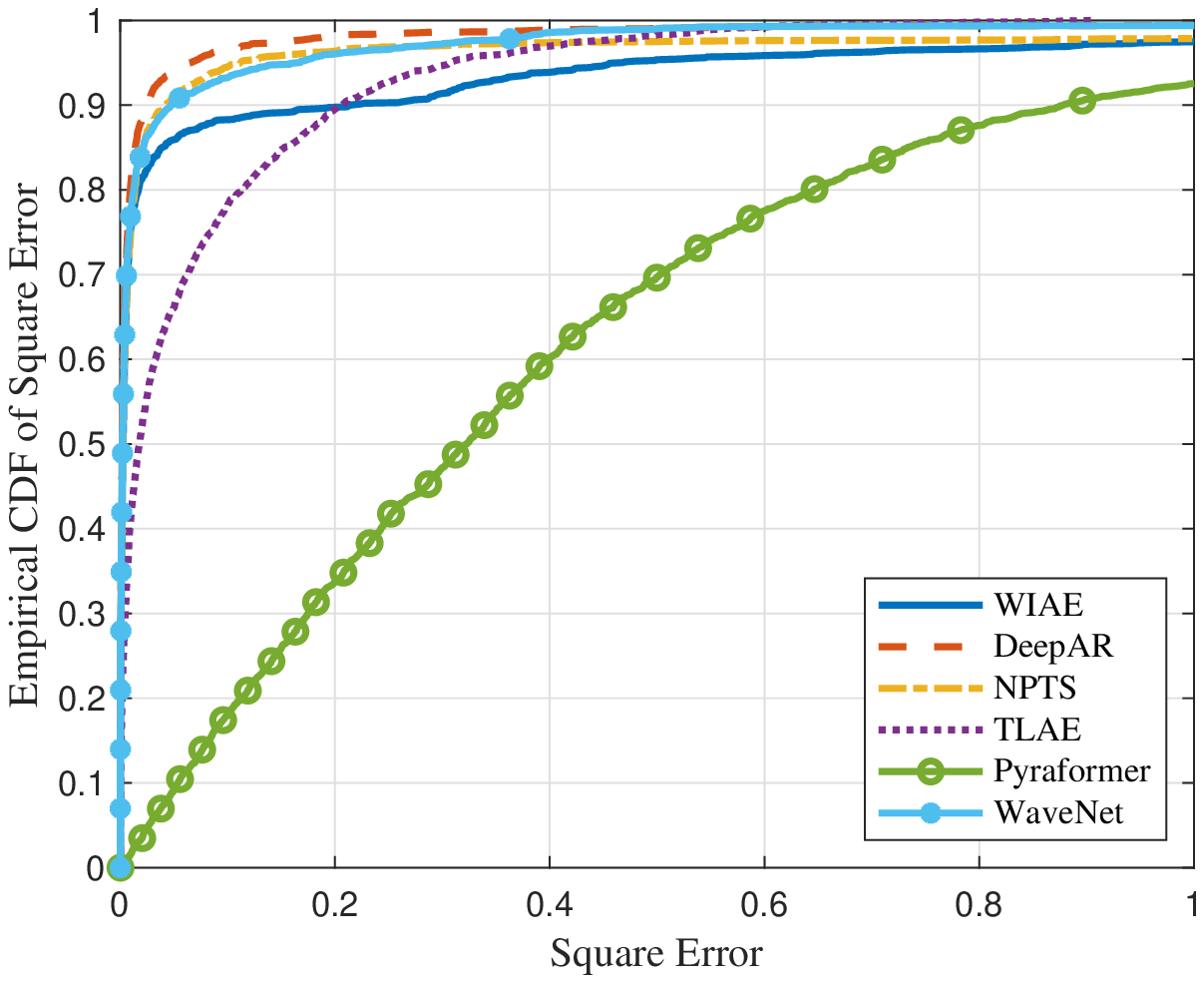}
    \caption{Errors for PJM 5-step (two subplots to the left) and 12-step (two subplots to the right) prediction.}
    \label{fig:ecdf_pjm}
\end{figure}

The metrics listed in Sec.~\ref{sec:metrics} are not able to comprehensively represent the performance of different methods, for mean and median are only two statistics that can be calculated from the error sequence of each method for different datasets. 
Especially for electricity price datasets with high variability, the distribution of error is quite skewed, which means mean and median errors can be quite different.
To provide a more comprehensive view of how different methods performed on electricity datasets, we plotted the empirical cumulative function (eCDF) of errors for each datasets.
The eCDF of an $N$-point error array (denoted by $F_N(\cdot)$) can be computed given:
\[F_N(t) = \frac{1}{N}\sum_{i=1}^N\mathbbm{1}_{e_i\leq t},\]
where $(e_i)$ is the error sequence, and $\mathbbm{1}$ the indicator function.
Hence, if the eCDF of a method is goes to $1$ when $t$ is small, then it means that the majority of the errors are small.

Fig.~\ref{fig:ecdf_isone}-\ref{fig:ecdf_pjm} showed the eCDF for absolute errors and square errors for ISONE, NYISO and PJM datasets.
It can be seen from the figures that for ISONE and NYISO, WIAE had the best overall performance for under most cases.
For PJM dataset, DeepAR and TLAE had slightly better performance.

\end{document}